\documentclass{article}
\usepackage{float}
\usepackage[utf8]{inputenc} 
\usepackage[T1]{fontenc}    
\usepackage{hyperref}       
\usepackage{url}            
\usepackage{bm}

\usepackage{booktabs}       
\usepackage{amsfonts}       
\usepackage{nicefrac}       
\usepackage{microtype}      
\usepackage{xcolor}         
\usepackage{amsmath,amssymb}
\usepackage{wrapfig}
\usepackage{stfloats}
\usepackage{graphicx}
\usepackage{authblk}
\usepackage{caption}
\usepackage{subcaption}
\usepackage{multirow}
\usepackage[square,numbers]{natbib}
\NewDocumentCommand{\codeword}{v}{%
\texttt{\textcolor{blue}{#1}}%
}

\DeclareMathOperator*{\argmax}{arg\,max}

\newcommand{\Rb}{\mathbb{R}}

\newcommand{\Pb}{\mathbb{P}}

\newcommand{\Fc}{\mathcal{F}}

\newcommand{\Nc}{\mathcal{N}}

\newcommand{\Zc}{\mathcal{Z}}

\title{Deep State Space Recurrent Neural Networks for Time Series Forecasting}

\author[1]{Hugo Inzirillo}
\affil[1]{%
  CREST-ENSAE, Institut Polytechnique de Paris}

\begin{document}

\maketitle
\begin{abstract}
We explore various neural network architectures for modeling the dynamics of the cryptocurrency market. Traditional linear models often fall short in accurately capturing the unique and complex dynamics of this market. In contrast, Deep Neural Networks (DNNs) have demonstrated considerable proficiency in time series forecasting. This papers introduces novel neural network framework that blend the principles of econometric state space models with the dynamic capabilities of Recurrent Neural Networks (RNNs). We propose state space models using Long Short Term Memory (LSTM), Gated Residual Units (GRU) and Temporal Kolmogorov-Arnold Networks (TKANs). According to the results, TKANs, inspired by Kolmogorov-Arnold Networks (KANs) and LSTM, demonstrate promising outcomes.

\end{abstract}

\section{Introduction}
Digital assets constitute the most disruptive innovations of the last decade in finance. 
The primary intention of blockchain development was not to create a new
currency, but to establish the principles of a functional decentralised cash payment system \cite{P2P}. The launch of bitcoin in 2008 \cite{bitcoin}  was the wake-up call for the development of other crypto-currencies.
Since the last decade, we have witnessed the birth of several thousand cryptocurrencies according to \href{https://coinmarketcap.com/}{CoinMarketCap}. Many still wonder if this is the emergence of a new asset class or just a bubble \cite{bitcoin_bubble,CHEAH201532}.As this market has rapidly grown, so has the interest of researchers in this asset class. The particular characteristics of such assets have led some researchers to study their behaviour, in particular through statistical analysis and stochastic models on their returns \cite{LOPEZMARTIN2022387,inzirillo2021dimensionality}.

\medskip

The digital asset market is young and has seen an exponentially rapid growth in recent years. Empirical studies show that this disruptive market has differentiated itself from other traditional financial markets by particular features: very high volatilities \cite{KATSIAMPA20173}, inverse leverage effects \cite{GARCIAJORCANO2020101300,LOPEZMARTIN2022387}, skewed distributions, high kurtosis, etc. Despite the increasing interest in cryptocurrencies, it remains complicated to model the dynamics of their financial returns, mainly due to regularly observed periods of high volatility and to alternating booms and bursts \cite{KATSIAMPA20173}. A statistical study of bitcoin closing price distribution and dynamics cannot be induced by naive linear models, suggesting the existence of several market regimes. \cite{CRETAROLA2020108831} proposed an estimation procedure of all parameters based on the conditional maximum likelihood approach \cite{andersen1970asymptotic,cretarola2021detecting} and on Hamilton filtering \cite{hamilton}. In addition, deep neural networks have increasingly been used for time series modeling, showing better results compared to more classical approaches \cite{Deep_State_Space_Models,Attentive_State-Space_Modeling,NIKOLAEV2019723}. Some researchers have adopted a state-space approach using neural networks \cite{kuo2018markov,ilhan2021markovian}. There exists a rapidly growing literature on digital assets and deep learning applications to financial time series, particularly neural network models for forecasting purposes.
The novelty of these assets sets them apart from conventional financial assets, particularly in the behaviour of their returns and volatility. These characteristics make the task of modeling them more complex. Linear econometric models, such as ARMA, find their limits due to a lack of flexibility and their difficulty in inducing certain empirical characteristics. The growing interest lies in the ability of neural network-based methods to approximate highly nonlinear functions. The focus on Recurrent Neural Networks (RNNs) for forecasting returns, volatilities, risk measures, and other quantitative metrics in finance and economics, is well-justified, due to their unique architecture capable of capturing time dependency effectively. RNNs are particularly well-suited to sequential data, which makes these models a relevant choice for time series analysis, which is common in financial markets. However, the characteristics of a time series may vary depending on the regime (boom or bust, e.g.) we are in. Here, we introduce innovative neural network architectures that merge the principles of classic econometric state space models with RNNs. This is achieved by implementing a hidden switching mechanism among multiple networks, where the transition probabilities vary over time and are influenced by certain observable covariates. In the next section, we recall the framework of regime switching models which will be used later to be confronted with deep learning models to estimate model parameters, including time varying transition probabilities. In section \ref{Deep State Space Models section}, we will specify some deep learning models. We will begin with some basic models and progressively incorporate more complexity, aiming to propose extensions and improvements to the current state-of-the-art models.

\begin{figure}[H]
    \centering
    \includegraphics[scale = 0.3]{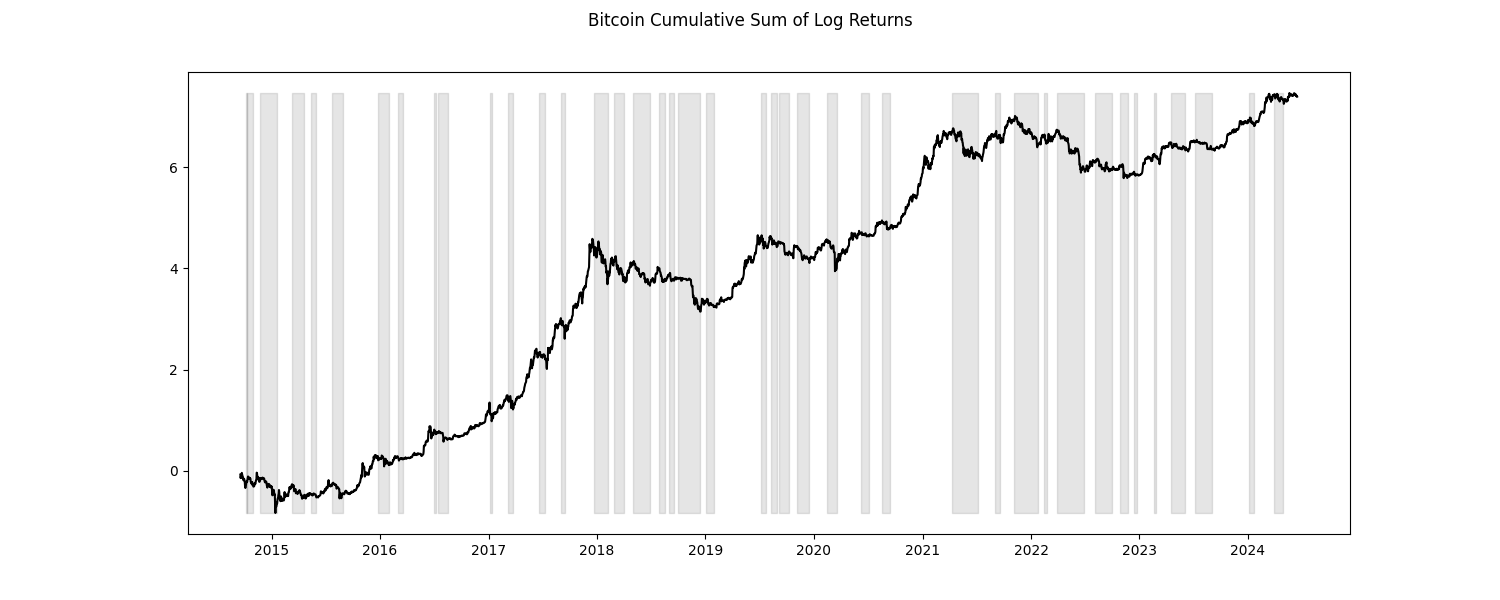}
    \caption{Bitcoin cumulative sum of log returns. The shaded part of the figure represents the downward trends observed. We have defined a "bearish regime" as the period when the 20-day rolling average of the cumulative sum of log returns, shifted by 20 days, is lower than the current 20-day rolling average of the cumulative sum of log returns.}
    \label{fig:bitcoin_prices}
\end{figure}

\section{Simple Regime Switching models with Time Varying Transition Probabilities}
\label{RegimeSwitching}
Regime switching models are widely used by econometricians to model nonlinear dynamics of financial returns. 
These models have been introduced by \cite{hamilton} in macroeconometrics.
These models are particularly relevant when certain time series exhibit distinct dynamics that depend on the state of the economy. We intend to apply the latter intuition to RNNs, particularly focusing on GRUs, LSTMs and TKANs. Our approach involves developing progressively more intricate models. We begin with GRUs, a specific type of RNN, and then enhance them by incorporating switching mechanisms. We will have the same approach for LSTMs and TKANs, introduced in the previous part of the thesis. Let us consider a (possibly non-homogeneous) Markov chain $(s_t)_{0{\leq t}{\leq T}}$. Here, $s_t \in \{1,\ldots,m\}$  may be interpreted as the state of the economy (or "the market") at time $t$. 
We will consider $(s_t)_{0{\leq t}{\leq n}}$ as an irreductible chain, with associated transition probabilities $ p_{ij,t}:=\mathbb{P}( s_t=j | s_{t-1}=i,\mathcal{F}_{t-1})$, for any $(i,j)$ in $\{1,\ldots,m\}$ and any time $t$. 
We will study a discrete time series of closing prices $(Y_t)_{0{\leq t}{\leq n}}$. 
The associated log-returns are $r_t:=\ln Y_t-\ln Y_{t-1}$. 
The series $(r_t)_{0{< t}{\leq n}}$ may most often be considered as strongly stationary for many financial assets (during reasonably long periods of time without any "structural break"), meaning that the joint distribution of the log returns denoted $(r_{t},..., r_{t-p})$ is independent of $t$ for all $p$. A basic regime switching model is typically written as
\begin{equation}
    r_t = \alpha_{s_t} + \epsilon_t, \quad \epsilon_t \sim \Nc(0,\sigma^2_{s_t}), 
\label{basic_SwitchReg}
\end{equation}
for some model parameters $(\alpha_1,\ldots,\alpha_m,\sigma_1,\ldots,\sigma_m)$.
Markov Switching models are useful to capture switches across market regimes. The seminal model of \cite{hamilton} assumed that the dynamics of the states $(s_t)$ is purely exogenous and independent of the realizations of the variables of interest $(r_t)$. 
 \cite{diebold_tvtp} extended the model by assuming that the transitions between regimes could be caused by some underlying explanatory variables. The full amount of information that is available at the beginning of time $t$ is denoted as $\mathcal{F}_{t-1}$.
Typically, $\mathcal{F}_{t-1}$ records all returns and explanatory variables that have been observed before time $t$. 
Then, this induces a filtration $\mathcal{F}:=(\mathcal{F}_{t})_{t\geq 0}$.
In a more general framework, we have to manage time varying transition probabilities across $m$ states, justifying the notation
$$
    p_{ij,t}:=\mathbb{P}( s_t=j | s_{t-1}=i,\mathcal{F}_{t-1}), \quad \sum_{j=1}^{m} p_{ij,t} = 1.
$$
We denote by $P_t$ the associated transition matrix for each time step $t \in \{0,\ldots,T\}$:
$$
P_{t} :=  \begin{pmatrix}
p_{11,t} & \dots & p_{1m,t}\\
 \vdots &  & \vdots\\
p_{m1,t} & \dots  & p_{mm,t}
\end{pmatrix}.
$$
Our transition matrix will be defined as a measurable map of some past covariates, i.e. $ P_t$ belongs to $\mathcal{F}_{t-1}$. 
To be specific, for any $i\in \{1,\ldots,m\}$, let $(z^{(i)}_t)_{t\geq 0}$ be a series of random vectors, where 
$z_{t-1}^{(i)}\in \mathcal{F}_{t-1}$ records the covariates that will be used to define the time varying transition probabilities from state $i$ between the times $t-1$ and $t$. 
For notational convenience, we concatenate the vectors $z^{(i)}_t$, $i\in \{1,\ldots,m\}$ to build a new vector $z_t$.
In terms of specification, for any $i,j\in \{1,\ldots,m\}$, $i\neq j$, and any $t$, we will set
\begin{equation}
\label{equation:transition_prob}
    p_{ij,t} :=\frac{\exp (\beta_{i,j} z_{t-1}^{(i)} )}{ \sum_{k=1}^{m} \exp (\beta_{i,k} z_{t-1}^{(i)}) } \cdot
\end{equation}
The latter time-varying transition probabilities depend on unknown (row) vectors of parameters $\beta_{i,j}$, $i,j\in \{1,\ldots,m\}$, $i\neq j$, that have to be estimated. The final vector of unknown parameters will be denoted as $ \theta$. It stacks the constants $\alpha_k$, $k\in \{1,\ldots,m\}$ and the slope parameters $\beta_{i,j}$, $(i,j)\in \{1,\ldots,m\}^2$, $i\neq j$. To estimate $\theta$, the maximum likelihood method is usually invoked. The associated log-likelihood function is given by
\begin{equation}
     L_T(\theta):=\sum_{t=1}^{T}  \log\big(f(r_t|{\mathcal F}_{t-1})\big),
\end{equation}
Note that, under (\ref{basic_SwitchReg}), we have
\begin{equation}
    f(r_t|s_t) = \frac{1}{\sqrt {2\pi \sigma_{s_t}^2} }{\exp \bigg(- \frac{(r_t - \alpha_{s_t})^2}{2\sigma_{s_t}^2}}\bigg).
\end{equation}
Note that the joint density of $r_t$ and the unobserved variable $s_t$ is
\begin{equation}
    f(r_t,s_t|z_{t-1}) = f(r_t|s_t,z_{t-1}) p(s_t|z_{t-1}),
\end{equation}
with obvious notations.
The marginal density of $r_t$ is obtained by summing over all the possible states:
\begin{equation}
\label{margin_density_rt}
    \begin{split}
        f(r_t|{\mathcal F}_{t-1}) & = \sum_{s_t=1} ^{m} f(r_t,s_t|z_{t-1}),\\
        & = \sum_{s_t=1} ^{m} f(r_t|s_t,z_{t-1}) p(s_t|z_{t-1}).\\
    \end{split}
\end{equation}
Using \eqref{margin_density_rt} we can rewrite the likelihood as
\begin{equation}
    L_T(\theta) :=\sum_{t=1}^{T}  \log\Big(\sum_{s_t=1} ^{m} f(r_t|s_t,z_{t-1}) p(s_t|z_{t-1})\Big).
\end{equation}
In the case of an autoregressive model of order $p$ with an underlying hidden first-order Markov chain  (called MSAR(p)), we would write the density of $r_t$ given the past information contained in $z_{t-1}$, we would need to consider $s_t$ as well as $s_{t-1}$. As mentionned before, $s_t$ is unobserved, hence to solve this issue we would consider the join density of $r_t$, $s_t$ and $s_{t-1}$.
\begin{equation}
        f(r_t,s_t,s_{t-1}|z_{t-1}) = f(r_t|s_t,s_{t-1},z_{t-1}) p(s_t,s_{t-1}|z_{t-1})
\end{equation}
$f(r_t|z_{t-1})$ can be computed by summing the possible values of $s_t$ and $s_t-1$ as follows: 
\begin{equation}
    \begin{split}
         f(r_t|z_{t-1}) & =\sum_{s_t}^{m} \sum_{s_{t-1}}^{m} f(r_t,s_t,s_{t-1}|z_{t-1}),\\
        & =\sum_{s_t}^{m} \sum_{s_{t-1}}^{m}f(r_t|s_t,s_{t-1},z_{t-1})\\p(s_t,s_{t-1}|z_{t-1})
    \end{split} 
\end{equation}
where $p(s_t,s_{t-1}|z_{t-1}) = p(s_t|s_{t-1},z_{t-1}) p(s_{t-1}|z_{t-1})$. To make prediction, one has to evaluate
$$ \hat r_t = E[r_t | \mathcal{F}_{t-1}]=\sum_{j=1}^{m} P(s_t=j| \mathcal{F}_{t-1})  E[r_t | \mathcal{F}_{t-1}, s_t=j]    .$$
\subsection{Simple Regime Switching (2 states)}
To proceed, we considered the daily log returns $r_t := \ln P_t- \ln P_{t-1}$ of Bitcoin. Initially, we estimate a Markov switching model with constant transition probabilities in order to determine whether there exist two or rather three underlying states. Next, we take the analysis further by including covariates and time varying transition probabilities in the model. This allows us to understand how these covariates influence the likelihood of switching between different states, as captured by the transition matrix. Furthermore, by analyzing our time-varying transition matrices, we can assess to what extent the model's behaviour remains persistent over time. You will find below the results of the estimations of Model (\ref{basic_SwitchReg}) with two market regimes. The results with three market regimes are available in the appendix \ref{regime_3}-\ref{regimeTVTP_3}.
 
 \begin{figure}[H]
    \centering
    \includegraphics[scale=0.4]{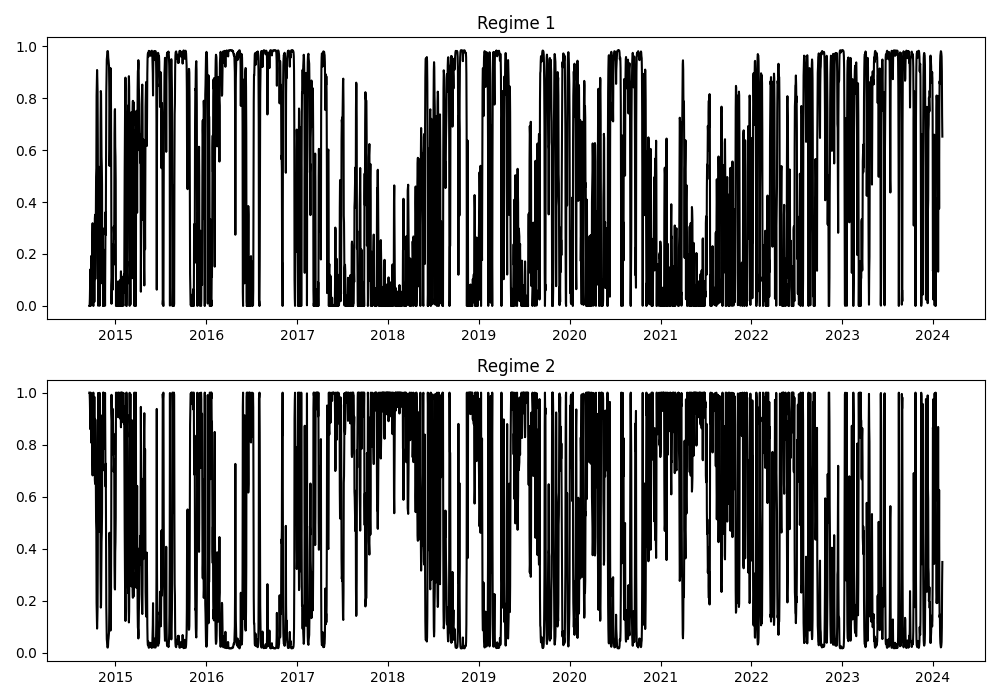}
    \caption{Smoothed Marginal Probabilities (basic MS model, 2 regimes)}
    \label{fig:2_regimes_model_without_tvtp}
\end{figure}

\begin{center}

\begin{table}[H]
    \centering

\begin{tabular}{lcccccc}
\toprule
                & \textbf{coef} & \textbf{std err} & \textbf{z} & \textbf{P$> |$z$|$} & \textbf{[0.025} & \textbf{0.975]}  \\
\midrule
\textbf{constant}  &       0.0014  &        0.001     &     2.677  &         0.007        &        0.000    &        0.003     \\
\textbf{variance} &       0.0002  &     5.15e-05     &     4.192  &         0.000        &        0.000    &        0.000     \\
                & \textbf{coef} & \textbf{std err} & \textbf{z} & \textbf{P$> |$z$|$} & \textbf{[0.025} & \textbf{0.975]}  \\
\midrule
\textbf{constant}  &       0.0015  &        0.001     &     1.107  &         0.268        &       -0.001    &        0.004     \\
\textbf{variance} &       0.0024  &        0.000     &    11.890  &         0.000        &        0.002    &        0.003     \\
                   & \textbf{coef} & \textbf{std err} & \textbf{z} & \textbf{P$> |$z$|$} & \textbf{[0.025} & \textbf{0.975]}  \\
\midrule
\textbf{p[1-$>$1]} &       0.8643  &        0.051     &    17.028  &         0.000        &        0.765    &        0.964     \\
\textbf{p[2-$>$1]} &       0.1479  &        0.049     &     3.033  &         0.002        &        0.052    &        0.244     \\
\bottomrule
\end{tabular}
\caption{Markov Switching Model Results (basic MS model, 2 regimes)}
    \label{tab:my_label_2_states_non_tvtp}
\end{table}
\end{center}
The table \ref{tab:my_label_2_states_non_tvtp} shows the model's estimated coefficients, their associated standard errors, and p-values for a two-regimes non-time-varying Markov switching. For instance, in state 1, the constant term's coefficient is noted as 0.0014, with a standard error of 0.001 and a p-value of 0.007, indicating a statistically significant difference from zero at the 1\% significance level. Similarly, the coefficient for the probability of remaining in state 1 is reported as 0.8643, with a standard error of 0.051 and a p-value practically equal to zero, suggesting a highly significant estimate that challenges the hypothesis of a coefficient value of 1 at the 0.1\% level. In state 2, the constant term coefficient is noted as 0.0015 with a p-value of 0.268, meaning the null hypothesis cannot be rejected: the constant term is not statistically different from zero. The relatively high values of $p[1 \to 1]$ and $p[2 \to 2]$ prove the tendency for the system to remain in its current state rather than rapidly transitioning to another one, which is a sign of stability of the model.
Finally, we observe a large difference in terms of conditional variances between regime 1 and regime 2 (0.0002 vs 0.0024, respectively). Cryptocurrency markets are then inherently volatiles and susceptible to break.

\medskip

The analysis of similar results in the case of three regimes (Figure \ref{regime_3}) show a different and more puzzling picture. The first two regimes are highly volatiles, with frequent switched between them. Due to low means and volatility, Regime 1 appears as rather artificial and spurious. The third is more clear-cut, since it is very persistent and exhibits a high level of volatility. Thus, in terms of interpretation and financial intuition, a two state MS model seems to be more realistic.
\subsection{Covariates}
For each models, we first estimate smooth marginal probabilities. In our analysis, we would consider two states, $s_t \in \{0,1\}$. Our prior is that one state is related to normal return and the other is related to a state with more agitation.
We assume :\\
\begin{equation}
    r_t=\begin{cases}
    a_1 +\sigma_{1}\epsilon_t, & \text{if $s_t=1$}.\\
    a_2 +\sigma_{2}\epsilon_t, & \text{if $s_t=2$}.\\
  \end{cases}
  \label{estimation_processus}
\end{equation}\\
Where $ a_2\geq a_1 $. We expect that $\sigma_2 > \sigma_1$. To compute the time varying transition probabilities, we used the following covariates: High Minus Low (HML) and an intraday variance indicator denoted $(IV)$ following the methodology of \cite{inzirillo2021dimensionality}. In this paper, they show the impact of such factor and also its capacity to sum up the information in a subspace using statistical techniques to reduce the dimension of the input data. We  define $\theta^{i}_t \in R^{4}_{+}$, the set of parameter for the i-th asset. $\theta^{i}_t:= \{O^{i}_t,H^{i}_t, L^{i}_t,C^{i}_t\}$. 
The proxy of intraday volatility of the asset is given by,  
\begin{equation}
\label{eq:iv}
    f_{t}(\theta^{i}_t):=  \Psi(H^{i}_t,L^{i}_t),
\end{equation}
where 
$$
    \Psi=\begin{cases}
    \log \left(\frac{H^{i}_t}{L^{i}_t}\right), & \text{if $O^{i}_t\geq C^{i}_t$}.\\
    \log \left(\frac{L^{i}_t}{H^{i}_t}\right), & \text{if $O^{i}_t<C^{i}_t$}.
  \end{cases}
$$
and we obtain,
\begin{equation}
    IV_{t}^{i} = f_{t}(\theta^{i}_t)^2
    \label{eq:iv}
\end{equation}
\textit{For deep learning tasks, if needed, after constructing our vector of covariates for each date $t$, we apply standardization on our dataset in order to proceed to the estimation of the model parameters for the different market regimes.}

\subsection{Impact of Covariates}
\begin{figure}[H]
    \centering
    \includegraphics[scale=0.4]{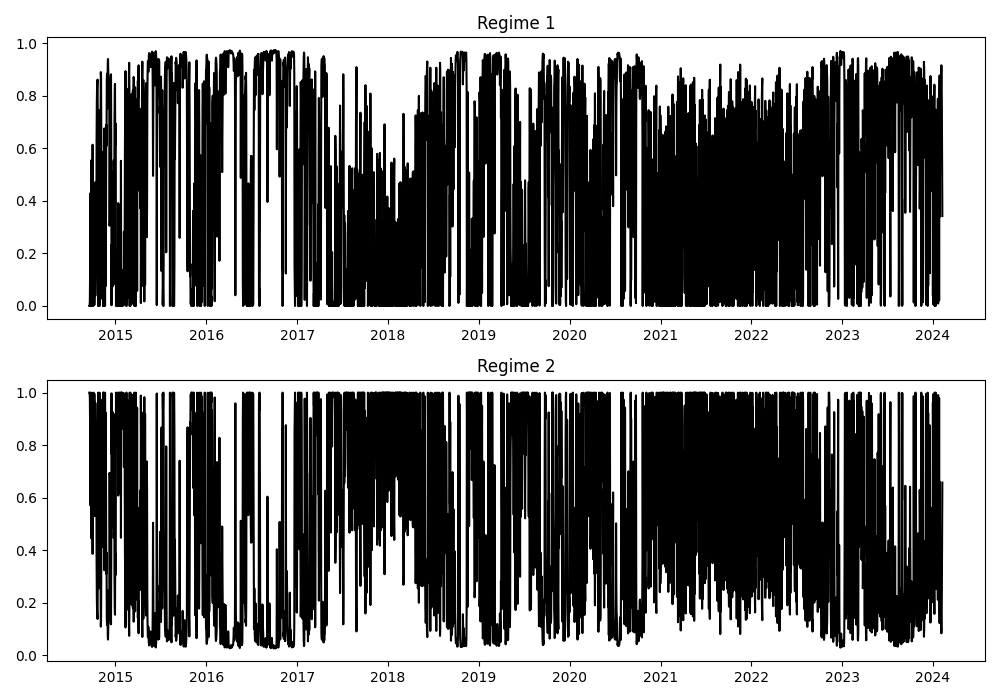}
    \caption{Smoothed Marginal Probabilities}
    \label{fig:2_regimes_model_with_tvtp_hml_vol}
\end{figure}

This section explores how additional informations (covariates) affect the likelihood of market regime changes. We are interested in seeing if these covariates help us identify different market regimes. Initially, we tested a model with two volatility regimes (a regime with high volatility and another one with low volatility), in terms of asset returns. First, we study time-varying transition probabilities using an ``High-Minus-Low'' (HML) indicator as a covariate. Subsequently, we extend our analysis by integrating other indicators, notably the "intraday volatility" (IV, see \eqref{eq:iv}), to assess their influence on the accuracy of our estimated parameters. Our goal is to analyze the dynamics of the time-varying transition matrices when such factors are used, and to verify whether their integration induces significant different interpretations. Table \ref{tab:2_regimes_model_with_tvtp_hml} displays the results using HML only. We clearly still identify the presence of two distinct regimes differentiated by their conditional variances. The introduction of HML inside the model as a covariate for the estimation of time varying transition probabilities makes sense. The new coresponding coefficients are statistically different from zero, indicating a real effect of HML on transition probabilities. When we enrich the model with IV (see Figure \ref{fig:2_regimes_model_with_tvtp_hml_vol} and Table \ref{tab:2_regimes_tvtp_with_hml_iv}),
the picture does not change, and we conclude that using HML and IV as drivers of time-varying transition probabilities is meaningful. The same experiment with three regimes (Table \ref{tab:2_regimes_tvtp_with_hml_iv}) provides a less clear-cut view, as without covariates. In particular, many estimated coefficients related to HML and/or IV are no longer statistically different from zero, casting doubt on model specification.
In view of the latter results, we preferred to consider two states for the model, where state 2 is related to relatively more volatile asset returns than state 1.

\subsubsection*{Regime Switching TVTP with HML Factor}

\begin{figure}[H]
    \centering
    \includegraphics[scale=0.4]{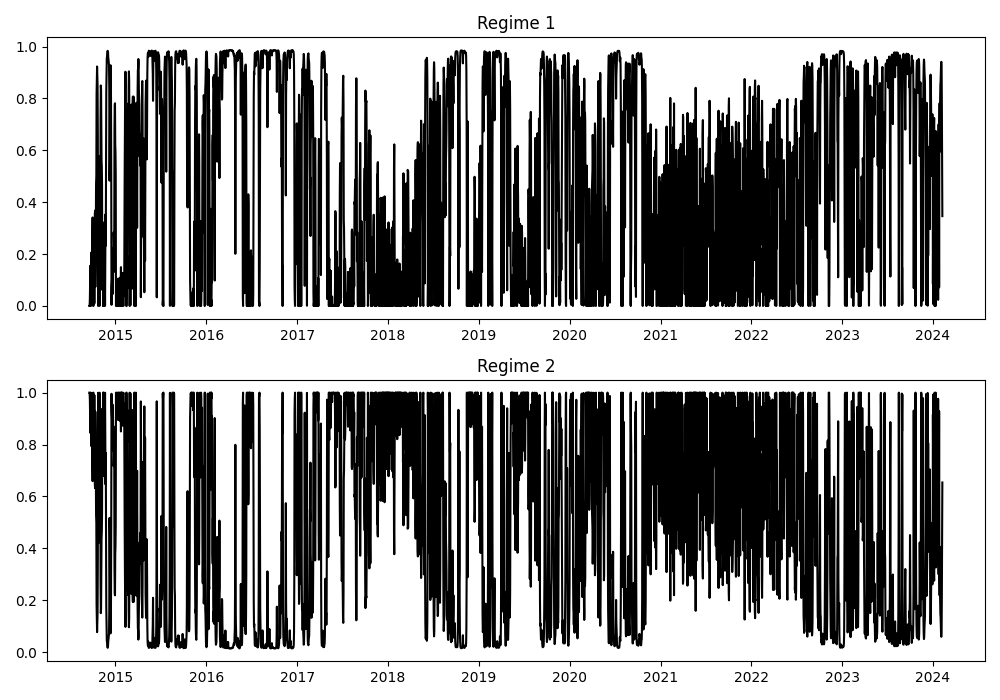}
    \caption{Smoothed Marginal Probabilities}
    \label{fig:2_regimes_model_with_tvtp_hml}
\end{figure}

\begin{table}[H]
    \centering
   \begin{center}
\begin{tabular}{lcccccc}
\toprule
                & \textbf{coef} & \textbf{std err} & \textbf{z} & \textbf{P$> |$z$|$} & \textbf{[0.025} & \textbf{0.975]}  \\
\midrule
\textbf{const}  &       0.0013  &        0.001     &     2.552  &         0.011        &        0.000    &        0.002     \\
\textbf{sigma2} &       0.0002  &     5.13e-05     &     3.817  &         0.000        &     9.53e-05    &        0.000     \\
                & \textbf{coef} & \textbf{std err} & \textbf{z} & \textbf{P$> |$z$|$} & \textbf{[0.025} & \textbf{0.975]}  \\
\midrule
\textbf{const}  &       0.0016  &        0.001     &     1.190  &         0.234        &       -0.001    &        0.004     \\
\textbf{sigma2} &       0.0025  &        0.000     &    10.552  &         0.000        &        0.002    &        0.003     \\
                         & \textbf{coef} & \textbf{std err} & \textbf{z} & \textbf{P$> |$z$|$} & \textbf{[0.025} & \textbf{0.975]}  \\
\midrule
\textbf{p[1-$>$1].const} &       1.2185  &        0.439     &     2.775  &         0.006        &        0.358    &        2.079     \\
\textbf{p[2-$>$1].const} &      -1.3992  &        0.258     &    -5.424  &         0.000        &       -1.905    &       -0.894     \\
\textbf{p[1-$>$1].hml} &      -1.2059  &        0.236     &    -5.115  &         0.000        &       -1.668    &       -0.744     \\
\textbf{p[2-$>$1].hml} &       0.2336  &        0.111     &     2.105  &         0.035        &        0.016    &        0.451     \\
\bottomrule
\end{tabular}

\end{center}
\caption{Markov Switching Model Results, Figure \ref{fig:2_regimes_model_with_tvtp_hml}}
\label{tab:2_regimes_model_with_tvtp_hml}
\end{table}

\subsubsection*{Regime Switching TVTP with HML and IV Factors}

\begin{table}[H]

\begin{center}

\begin{tabular}{lcccccc}
\toprule
                & \textbf{coef} & \textbf{std err} & \textbf{z} & \textbf{P$> |$z$|$} & \textbf{[0.025} & \textbf{0.975]}  \\
\midrule
\textbf{const}  &       0.0012  &        0.000     &     2.777  &         0.005        &        0.000    &        0.002     \\
\textbf{sigma2} &       0.0002  &     2.67e-05     &     6.553  &         0.000        &        0.000    &        0.000     \\
                & \textbf{coef} & \textbf{std err} & \textbf{z} & \textbf{P$> |$z$|$} & \textbf{[0.025} & \textbf{0.975]}  \\
\midrule
\textbf{const}  &       0.0017  &        0.001     &     1.353  &         0.176        &       -0.001    &        0.004     \\
\textbf{sigma2} &       0.0025  &        0.000     &    13.731  &         0.000        &        0.002    &        0.003     \\
                         & \textbf{coef} & \textbf{std err} & \textbf{z} & \textbf{P$> |$z$|$} & \textbf{[0.025} & \textbf{0.975]}  \\
\midrule
\textbf{p[0-$>$0].const} &      -0.1682  &        0.456     &    -0.369  &         0.712        &       -1.062    &        0.726     \\
\textbf{p[1-$>$0].const} &      -0.3797  &        0.259     &    -1.467  &         0.142        &       -0.887    &        0.128     \\
\textbf{p[0-$>$0].hml} &      -0.8880  &        0.312     &    -2.848  &         0.004        &       -1.499    &       -0.277     \\
\textbf{p[1-$>$0].hml} &       0.4328  &        0.146     &     2.956  &         0.003        &        0.146    &        0.720     \\
\textbf{p[0-$>$0].iv} &      -1.5306  &        0.502     &    -3.049  &         0.002        &       -2.514    &       -0.547     \\
\textbf{p[1-$>$0].iv} &      -0.8256  &        0.192     &    -4.304  &         0.000        &       -1.202    &       -0.450     \\
\bottomrule
\end{tabular}
\end{center}

\caption{Regime Switching Parameters (HML,IV), Figure \ref{fig:2_regimes_model_with_tvtp_hml_vol} }
\label{tab:2_regimes_tvtp_with_hml_iv}
\end{table}

\newpage

\section{The Evolution of Deep Learning Methods}
  In this section, we review the fundamental architectures of neural networks, from the first basic attempts to the most advanced models as documented in the literature. Afterwards, we introduce our innovative state space neural network architecture, with a new switching mechanism. We also proposed a new type of layer for neural networks combining the power of RNNs with the efficiency of attention-free transformer architecture called the AF-LSTM layer.

    \subsection{Feed Forward Networks (FNNs)}
        The first stage of the story is based on the concept of MP Neuron \cite{mcculloch43a}. 
    Such MP neurons were originally used for classification tasks. They take binary inputs and return a binary output according to a trigger.

     \begin{figure}[H]
     \centering
     \includegraphics[width=10.0cm]{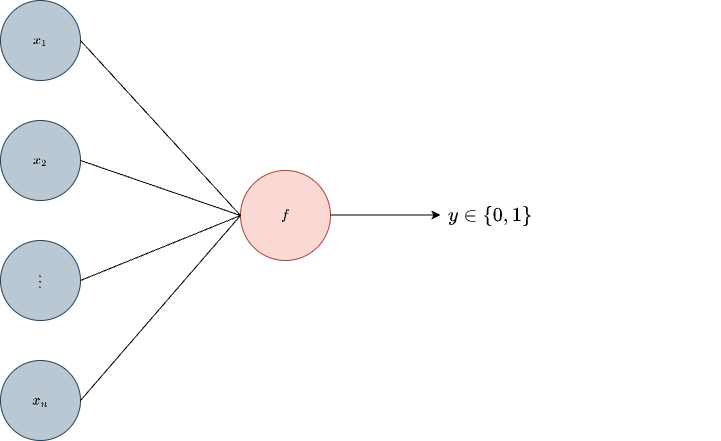}
     \caption{MP neuron}
     \label{fig:mp}
     \end{figure}
    
     Figure \ref{fig:mp} shows the structure of a basic MP neuron, composed by an input vector $X$ and a function $f$ which takes two values 0 or 1. 
     If the sum of the $x_i$, $ i \in \{1,\ldots,n\}$, is higher than a trigger $b$, then the value of the ouput $y$ is set to 1, and 0 otherwise. Mathematically, we can write the MP neuron predictor as follows:
    
     $$
     f(x) := \begin{cases}
             1 & \text{if } \sum_{i=1}^n x_i  \geq b .\\
             0   & \text{otherwise}.
       \end{cases}
     $$

    \subsubsection*{Perceptron}

    Perceptrons, introduced by \cite{rosenblatt1958perceptron} are an extensions of the previous MP neurons \cite{mcculloch43a} that can take any real value as input. Now, each element of the input vector $X$ is multiplied by a weight. Moreover, contrary to MP neurons, one or several layers are usually added in the network.
    This significantly increases the flexibility of the model.
    The addition of a vector of parameters $\theta$ introduced the error-correction learning in the neural network and made possible to adjust the weights to improve the classification task.

    \begin{figure}[H]
    \centering
    \includegraphics[width=10.0cm]{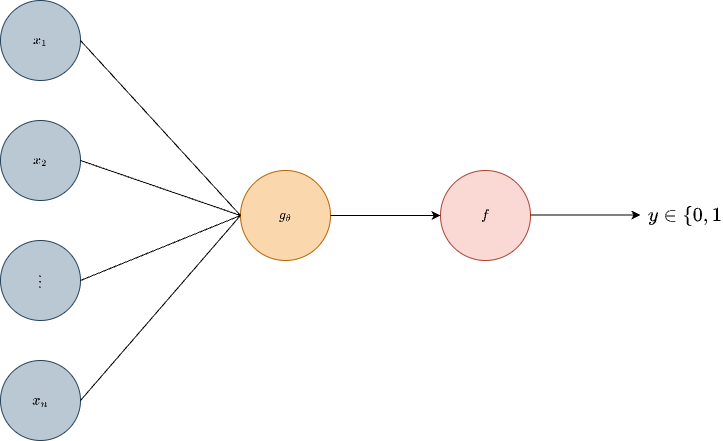}
    \caption{Perceptron}
    \label{fig:perceptron}
    \end{figure}

    In Figure \ref{fig:perceptron}, $g_\theta$ is a parametrized function which takes an input $\{x_i\}_{i=1}^{n}$ where $x_i$ is real number for every $i$. In a second stage, the output of $g_\theta$ will be given as inputs of $f \in F$, where $F$ is a set of "activation functions" which yields the final decision and return a binary variable (in the case of classification). Mathematically, this can be written as follows:
    
    $$
    f\big(g_{\theta}(x)\big) := \begin{cases}
            1 & \text{if } g_{\theta}(x) \geq b .\\
            0   & \text{otherwise}.
      \end{cases}
    $$
    Note that the threshold $b$ is a learning parameter that has to be set during the training stage of the network. Despite the power of MP neurons and perceptrons, they cannot easily manage non linearity in problem solving. Although Perceptrons have an embedded layer, their amount of flexibility is limited. To solve more complex nonlinear problems, Multi Layer Percetron (MLP) have been introduced. These arthictectures are a cornerstone of modern neural network research and applications.
    \subsubsection*{Multi Layer Neural Networks}
     \begin{figure}
    \centering
    \includegraphics[width=12.0cm]{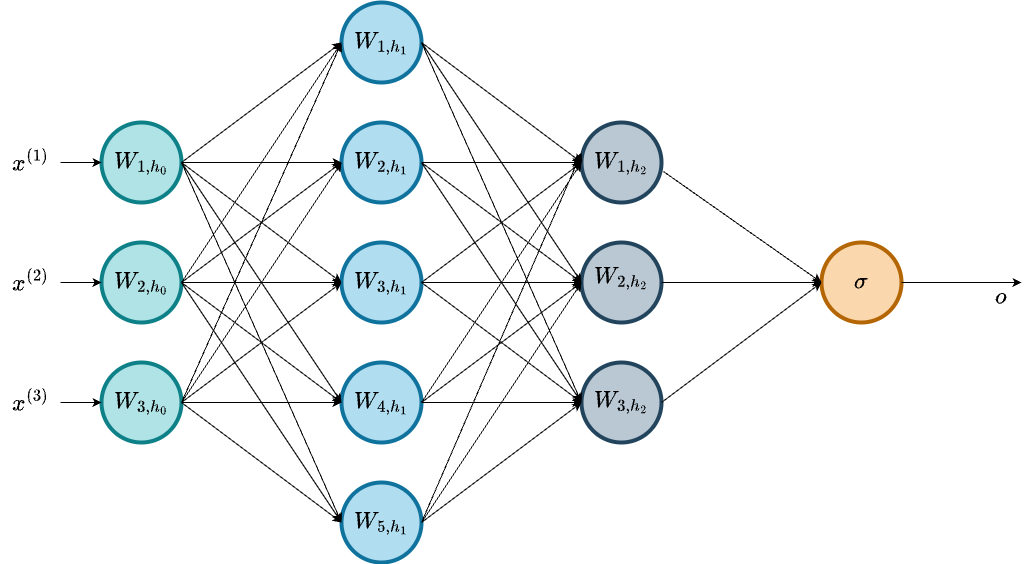}
    \caption{MLP}
    \label{fig:MLP}
    \end{figure}
    Let us define some notations: $W_{1}$ denotes the weights of the input layer, the first layer of the network. $W_{h_{p}}, \text{and} \; p \in [0,...,n] $ denotes the weight of the hidden layers and $n-1$ the number of total hidden layers. $W_{h_{n}}$ the vector of weight of the final hidden layers is then ouput layer of the network. $\sigma (\cdot)$ is the activation function of the network. In this case, we will use the logistic function:
    $$
        \sigma (z) = \frac{1}{1+\exp(-z)},
    $$
    step after step calculation is given by the following equations,
    \begin{equation}
        \begin{split}
        h_1 & = \sigma(W_{h_0}x+ b_{h_0}),\\
        h_2 & = \sigma( W_{h_2}h_1+ b_{h_1}),\\
        o  & = \sigma (W_{h_3}h_2+b_{h_2}). \\
    \end{split}
    \end{equation}

    As observed in Figure \ref{fig:MLP}, an MLP \cite{hornik1989multilayer,haykin1998neural,cybenko1989approximation} is an acyclic graph, oriented in one direction (from left to right, here). The layers are fully connected to each others. However, there is no connection between nodes within a layer. FFNs process information layer by layer without feedback connections. The missing feedback connections make them unsuitable for sequential data. FFNs process each element of the input independently, treating it as an isolated piece of information. These networks are unable to consider context or previous values, which are essential for understanding sequential data such time series. Nonetheless, perceptrons have yielded basic building blocks for a lot more complex and flexible predictor such recurrent neural networks.

    \subsection{Recurring Neural Networks (RNNs)}
    
\begin{figure}[H]
    \centering
    \includegraphics[width=10.0cm]{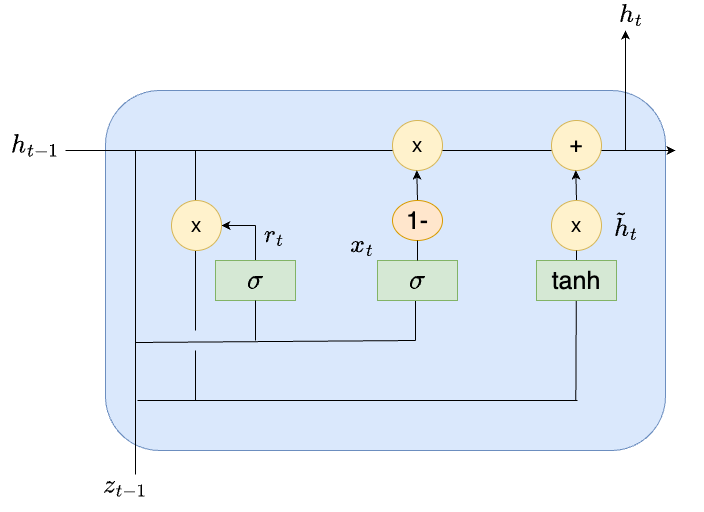}
    \caption{GRU}
    \label{fig:GRU}
    \end{figure}

    Recurrent Neural Networks (RNNs) have been proposed to address the "persistence problem", i.e. the potentially long-term dependencies between the successive observations of some time series. Here, an iterative process inside every cell allows information to persist through time and brings consistency with respect to temporal dependency. Therefore, RNNs most often outperform "static" networks as MLPs \cite{le2015simple}. Gated Recurrent Units (GRU) (Figure \ref{fig:GRU}) were proposed by \cite{cho_neural_machine}. They consitute the building blocks of some family of RNNs that explicitely take into account time ordering. A GRU embeds two "gated" mechanisms called "reset" and "update" detailed equations \eqref{eq:gru}.

    \medskip
    
    \begin{equation}
    \begin{split}
        x_t &= \sigma(W_z[h_{t-1},z_{t-1}])\\
        r_t &= \sigma(W_r[h_{t-1},z_{t-1}])\\
        \tilde{h}_t &= \text{tanh}(W[r_t \odot h_{t-1}, z_{t-1}])\\
        h_t &= (1-a_t) \odot h_{t-1} + z_t \odot \tilde{h}_t
    \end{split}
    \label{eq:gru}
    \end{equation}
    Traditional methods of gradient descent may not be sufficiently effective for training Recurrent Neural Networks (RNNs), particularly in capturing long-term dependencies \cite{learning_lt}. Meanwhile, \cite{grn_evaluation} conducted an empirical study revealing the effectiveness of gated mechanisms in enhancing the learning capabilities of RNNs. Actually, RNNs have proved to be one of the most powerful tools for processing sequential data and solving a wide range of difficult problems in the fields of automatic natural language processing, translation, image processing and time series analysis. Researchers have been able to mimic human abilities like selectively focusing on crucial information, similar to how our attention works on input sequences. This innovative mechanism will be discussed in a later section.

    \subsubsection*{Long-Short Term Memory (LSTM)}
    
    \begin{figure}[H]
    \centering
    \includegraphics[width=12.0cm]{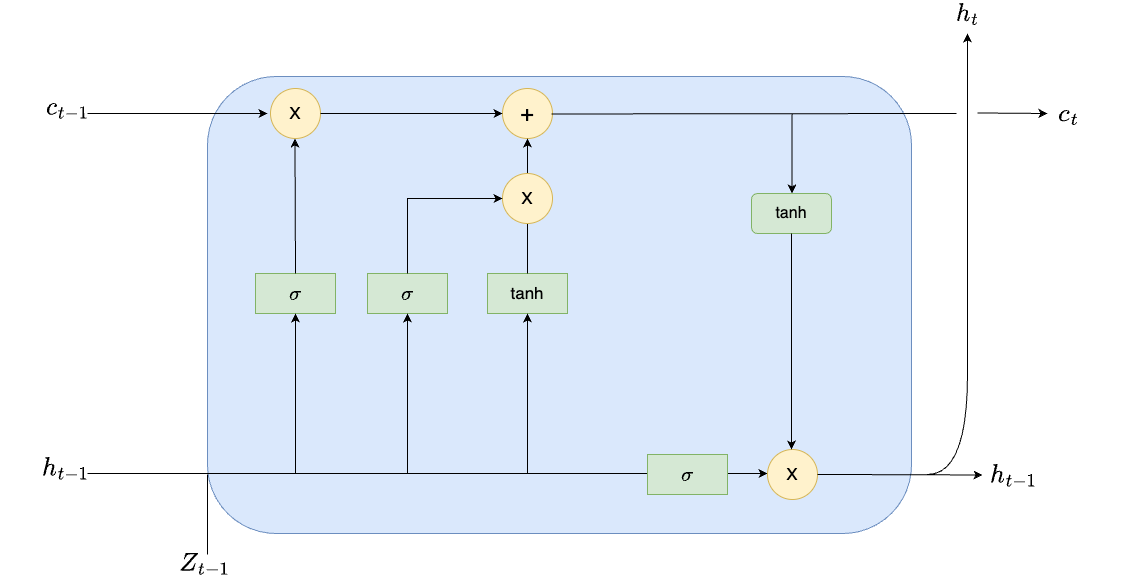}
    \caption{LSTM Cell}
    \label{fig:LSTM-Cell}
    \end{figure}
    
    Long-Short Term Memory (LSTM) networks \cite{CHARLES2019179} are specific type of RNNs with gated mechanisms. Like GRUs, LSTMs seek to control the flow of information through some gates without having to use a memory unit. However, GRUs have only two gated mechanisms whereas LSTM cell has three: the input gate, forget gate and update gate. One key particularity of LSTM architecture is that the update and forget gates are separated, which makes LSTMs more complex and evolved than GRUs. They have been designed to avoid the "vanishing gradient" problem. The latter problem often appears during the update of the usual RNN model proportionally weighted to the loss partial derivatives. Sometimes, the gradients of error terms may be vanishingly small and weights may not be updated during the learning task.
    To be specific, the LSTM cell built on Figure \ref{fig:LSTM-Cell} is defined by the following equations:
    \begin{equation}
    \label{eq:lstm}
    \begin{split}
        f_{t} & = \sigma(W_{f}[h_{t-1},z_{t-1}]+ b_{f})\\
        i_{t} & = \sigma(W_{i}[h_{t-1},z_{t-1}]+ b_{i})\\
        {\tilde c}_{t} & = {\tanh (W_{c}[h_{t-1},z_{t-1}]+ b_{c})}\\
        c_{t} & = f_{t} \odot c_{t-1}+i_{t} \odot {\tilde c}_{t}\\
        o_{t} & = \sigma(W_{o}[h_{t-1},z_{t-1}]+ b_{o})\\ 
        h_{t} & = o_{t} \odot {\tanh(c_{t})}, 
    \end{split}
    \end{equation}
    where $i_t,f_t$ and $o_t$ denote the input, forget and output gate, respectively. Our set of parameters, denoted by $ \theta := \{W_f, W_i, W_c, W_o,b_f,b_i,b_c,b_o\}$, stacks the weights and intercepts of the model and  $z_{t-1} \in \Rb^{s_{len} \times d}$ is the input vector at time $t$. $s_{len}$ denotes the lenght of the input sequence and $d$ the number of features. The notation  $g(z_{t-1};\theta):= \text{LSTM}(z_{t-1};\theta)$ represents the sequence of operations performed by the LSTM on $z_{t-1}$ with parameters $\theta$. In the case of financial time series prediction, we seek to predict future returns, volatilities, etc., based on the history of past returns. For instance, $\mathbb{E}[r_{t}|\mathcal{F}_{t-1}]$ would be estimated by a linear transformation of the LSTM outputs, i.e. by $\hat y_t:= W_y g(z_{t-1};\theta) + b_y$, for some parameters $W_y$ and $b_y$.

\section{Deep State Space Models}
\label{Deep State Space Models section}

\subsection{Switching Mechanism}
\label{sec:switching_mechanism}
In this subsection, we propose a novel approach to estimate switching probabilities through neural networks. Let us consider a hidden Markov chain $(s_t)_{0{\leq t}{\leq T}}$.
Here, $s_t \in \{1,\ldots,m\}$  may be interpreted as the market regime at time $t$.  We will consider $(s_t)_{0{\leq t}{\leq n}}$, an irreductible chain, with the associated conditional transition probabilities $ \Pb( s_t=j | s_{t-1}=i,\mathcal{F}_{t-1})$, for any $(i,j)$ in $\{1,\ldots,m\}$ and any time $t$ and for some user-defined filtration $(\mathcal{F}_{t})_{t\geq 0}$. \cite{hamilton} assumed that the dynamics of the states $(s_t)$ is purely exogenous and independent of the realizations of the variables of interest $r_t:=\log(p_t)-\log(p_{t-1})$.  \cite{diebold_tvtp} extended the model by assuming that the shifts between different states or regimes may be influenced by some underlying factors or explanatory variables. By default, $\mathcal{F}_{t-1}$ denotes the whole set of information accessible at the start of time 
$t$ (asset returns, volumes, general purpose market information, etc). 
The proposed switching mechanism allows us to estimate the conditional probability of being in a given state for each time step $t$
\begin{equation}
   \pi_{i,t|t-1} = \Pb( s_t=i | \Fc_{t-1}) \quad i \in \{1,...,m\},
   \label{red_pit}
\end{equation}
where $m$ denotes the total number of market regimes. First, introduce 
the $\sigma$-algebra induced by a time series of random vectors $(\Zc_j)_{j\leq t}$, i.e. 
$\Fc_{\Zc,t-1}=\sigma(\Zc_t, \Zc_{t-1},\ldots)$, and the associated filtration $(\Fc_{\Zc,t})_{t\geq 0}$.  
Typically, $\Zc_t$ will be the vector obtained from a neural network that will be built from some 
financial time series (including quotes, volumes, bid-ask spreads, or other market information possibly) until and including $t-1$.
In particular, $\Zc_{t}$ is $\Fc_{t-1}$-mesurable.
We now assume that $\Fc_{\Zc,t-1}$ brings a sufficient information to evaluate the conditional probabilities $\pi_{i,t|t-1}$, i.e., 
\begin{equation}
   \pi_{i,t|t-1} = \Pb( s_t=i | \Fc_{\Zc,t-1}), \quad i \in \{1,...,m\}.
   \label{red_pit}
\end{equation}
The latter probabilities $\pi_{i,t|t-1}$ will be estimated, computed recursively and updated at each time step using the 
(conditional) transition probabilities 
\begin{equation}
      \begin{split}
          p_{ij,t}&=\mathbb{P}( s_t=j | s_{t-1}=i,\Fc_{\Zc,t-1}).
      \end{split}
\end{equation}
To be specific, we will focus on the particular specification
\begin{equation}
    \begin{split}
        \Zc_t &= W_\Zc o_t  + b_\Zc, \\
    \end{split}
\end{equation}
where $W_\Zc \in \Rb^{m \times o_{dim}}$ and $o_t \in \Rb^{o_{dim}}$. Here, $o_t:=NN({\nu_{t-1}})$ denotes the output of a neural network. The ReLU (Rectified Linear Unit) activation function applied to the ouput of the neural network helps to filter the important information extracted from market information. The input of the neural networks $\nu_{t-1} \in \Rb^d$ is a stacked vector of some "covariates" that are observable at $t-1$. 
Among the latter variables, we have not considered the past returns: in the current model, the transition probabilities are influenced by the behaviour of covariates only. For each $t \in \{1,...,T\}$, we compute our transition probabilities to build our transition matrix
\begin{equation}
    P_{t} :=  \begin{pmatrix}
p_{11,t} & \dots & p_{1m,t}\\
 \vdots &  & \vdots\\
p_{m1,t} & \dots  & p_{mm,t}
\end{pmatrix}.
\end{equation}
At time $t$, our time varying transition probability matrix 
$ P_{t} =  P_{t-1} \rho_{t}$
provides a way of updating the transition matrix with some pieces of past information. 
To update the transition matrix $P_t$, we propose here to assume
\begin{equation}
    \rho_t = \exp(\Zc_t),
\end{equation}
where the exponential map is applied componentwise. $\rho_t$ will be subject to the transformation from a 1D vector to a 2D square matrix where rows and columns is associated to a market regime. Moreover, we can impose that the diagonal elements of $\rho_t$ are one. 

A true transition matrix $P_t$ is obtained from  
\begin{equation}
    P_t = \text{softmax}\big( P_{t-1} \odot \rho_t \big),
\end{equation}
where $\odot$ denotes componentwise multiplication. The SoftMax activation function is applied on every row of $P_{t-1} \odot \rho_t$. 

\medskip

As a second approach, it is tempting to enrich the latter way of estimating the latent states given some market information. Indeed, restricting ourselves to some "covariates" only may be questionable. Thus, we would like to add more information in the previous conditioning $\sigma$-algebra $\Fc_{\Zc,t-1}$.
Typically, at time $t$, having a value of the $t$-th return (or the $t$-volume, etc.) has to improve the prediction of $s_t$.
We particularize the additional stream of information that is induced by a sequence of random vectors $(y_t)_{t\geq 0}$.
At the beginning of time $t$, the available information $\Fc_{t-1}$ includes all $\Zc_{t-k}$, $k\geq 0$, all $y_{t-j}$, $j\geq 1$, and possibly other past market information. The new quantity of interest is denoted $\pi_{i,t|t}$ for each time step, with $ \pi_{i,t|t}:= \Pb( s_{t}=i | \Fc_{t})$.  
The new conditional probabilities $ \pi_{i,t|t}$ will be calculated by applying a type of Hamilton filter. Indeed, denoting $ \tilde\pi_{i,t|t-1}:= \Pb( s_{t}=i | \Fc_{t-1})$, we have
\begin{eqnarray}
\lefteqn{  \pi_{i,t|t} = \Pb( s_{t}=i | \Fc_{t}) \simeq  \frac{\Pb( s_{t}=i, y_t | \Fc_{t-1}) }{f(y_t|  \Fc_{t-1} ) }  }\nonumber \\ 
&=&  \frac{f( y_t | s_{t}=i, \Fc_{t-1}) \Pb( s_{t}=i |\Fc_{t-1}) }{\sum_{k=1}^m f(y_t|  s_{t}=k,\Fc_{t-1} ) \Pb( s_{t}=k |\Fc_{t-1})  }  \nonumber \\ 
&=&  \frac{f( y_t | s_{t}=i, \Fc_{t-1}) \tilde \pi_{i,t|t-1}}{\sum_{k=1}^m f(y_t|  s_{t}=k,\Fc_{t-1} ) \tilde \pi_{k,t|t-1}  } \cdot 
\label{rec_HF}
\end{eqnarray}
To justify the first identity, we implicitly assumed that $y_t$ is the single additional piece of information between $t-1$ and $t$, for the purpose of state forecasting. In particular, $\Zc_{t+1}$ does not matter.
Moreover, assume that $\Pb( s_{t}=j |s_{t-1}=i,\Fc_{t-1})=\Pb( s_{t}=j |s_{t-1}=i,\Fc_{\Zc,t-1})=p_{ij,t-1}$.
Thus, this yields
\begin{equation}
 \tilde \pi_{k,t|t-1} = \sum_{l=1}^m p_{kl,t-1} \pi_{l,t-1|t-1}. 
 \label{rec_pitt-1}
\end{equation}
Moreover, $  f( y_t | s_{t}=i, \Fc_{t-1})  $ can be evaluated as 
$$ 
 f( y_t | s_{t}=i, \Fc_{t-1}) = 
 \phi\big((y_t - \hat y_{i,t})/\sigma_{i,t}\big), 
$$
where $\phi$ is the density of a $\Nc(0,1)$. Here, every quantity $\hat y_{i,1}$, $i\in \{1,\ldots,m\}$, 
is an estimator of the conditional expectation of $y_t$ given its state. The latter quantities have been obtained by some neural networks.
The quantity $\sigma_{i,t}$ is the standard deviation of the quantities $\hat y_{i,1},\ldots,\hat y_{i,t-1}$.
In other words, we assumed the law of the explained variable $y_t$ is Gaussian, given its current state. 
This yields
\begin{equation}
\pi_{i,t|t} = \Pb( s_{t}=i | \Fc_{t}) =  
\frac{f( y_t | s_{t}=i, \Fc_{t-1}) \sum_{l=1}^m p_{kl,t-1} \pi_{l,t-1|t-1} }{\sum_{k,l=1}^m f(y_t|  s_{t}=k,\Fc_{t-1} ) p_{kl,t-1} \pi_{l,t-1|t-1}  } \cdot 
\label{rec_HF_2}
\end{equation}  
Finally, (\ref{rec_HF_2}) allows to recursively calculate the quantities $\tilde \pi_{i,t|t}$ and then the $\pi_{i,t|t-1}$ by (\ref{rec_pitt-1}).

\medskip

The quantities $\hat{y}_{i,t}$ refers to output of Neural Networks (NNs) within the Switching NN architecture will produce probabilities as an ouput stored in $\pi_t\in R^{m}$, for each time step. 

\medskip

\textit{During the learning task, detailed in the next section, we will apply this methodology on different recurrent neural networks. The objective is to assess the capacity of prediction by drawing a backtest to evaluate which models perform the best.}

\subsection{Switching RNNs}
\begin{figure}[H]
\centering
\includegraphics[width=12.0cm]{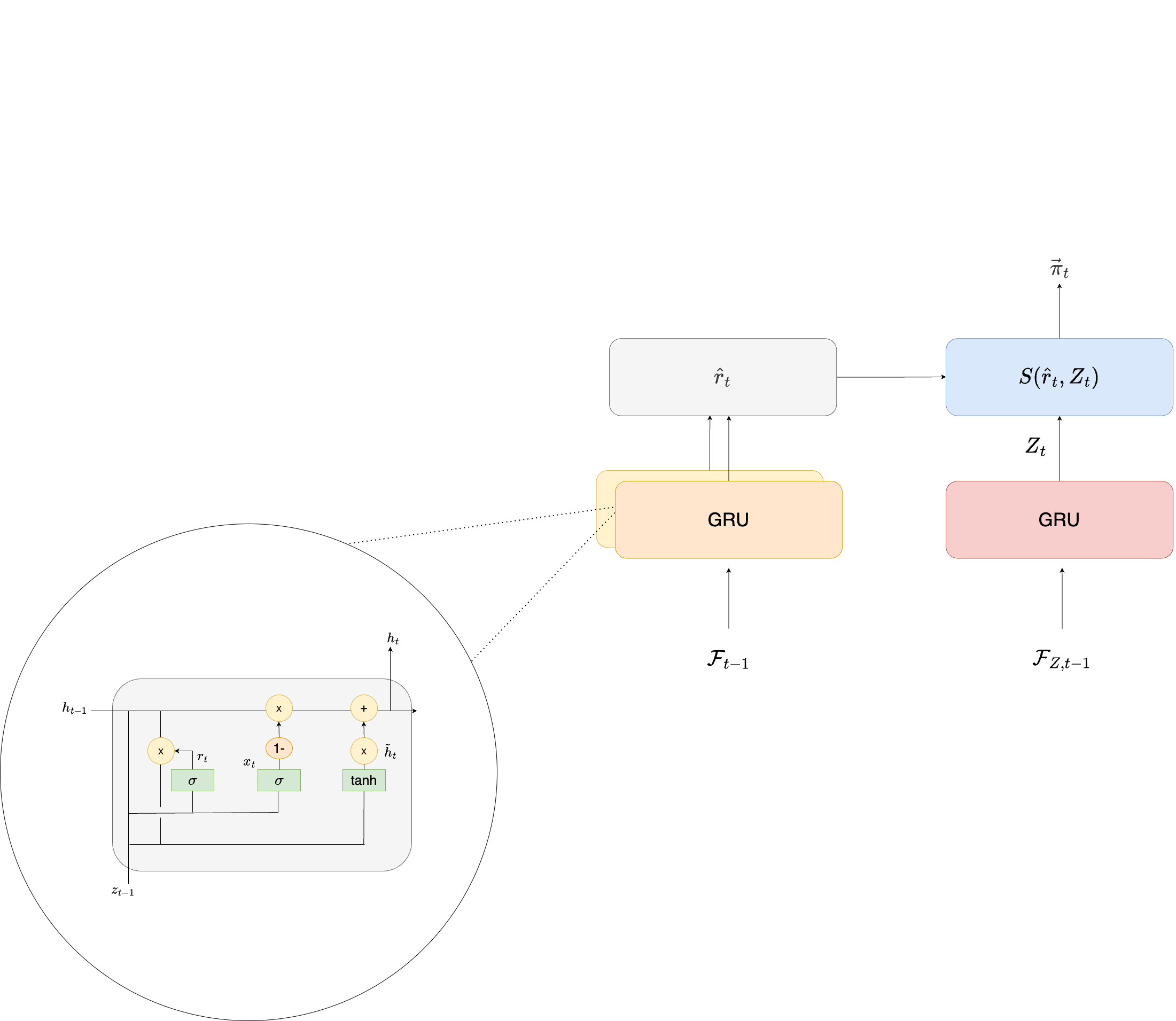}
\caption{Structure of m-GRU see \eqref{eq:k_lstm}}
\label{m-GRU}
\end{figure}
In the previous section, we have introduced the switching mechanism used to estimate transition probabilities and therefore the probability of being in a given state. Now, we will explore its application to several architectures. We will start with GRUs, known for its ability to efficiently capture long-term dependencies in sequences. We will see how this mechanism can help the GRU to better adapt to the different dynamics present in the covariates. Secondly, we will apply this mechanism to LSTMs known for their ability to manage short and long-term memory. Finally, we will apply this mechanism to the recently introduced TKAN architecture, which combines the use of KANs \cite{liu2024kan} and memory management. The Recurrent Neural Networks (RNNs) are a specific type of neural networks architecture oriented along a temporal sequence. This network architecture is distinguished from others by the presence of a memory effect, allowing it to process sequential data effectively.
One popular RNN variant is the Gated Recurrent Unit (GRU), known for its ability to capture long-term dependencies in sequences. The Switching GRU extends the standard GRU by taking into account a regime variable $k$ in the update, reset and hidden state gate equations:

\begin{equation}
\label{eq:k_gru}
\begin{split}
    z_t^{(k)} &= \sigma(W_{k,z} \cdot [h_{k,t-1}, z_{t-1}] + b_{k,z}),\\
    r_{k,t} &= \sigma(W_{k,r} \cdot [h_{k,t-1}, z_{t-1}] + b_{k,r}),\\
    \tilde{h}_{k,t} &= \text{tanh}(W[r_{k,t} \odot h_{k,t-1}, z_{k,t-1}]),\\
    h_{k,t} &= (1 - z_{k,t}) \odot h_{k,t-1} + z_{k,t} \odot \tilde{h}_{k,t}.
\end{split}
\end{equation}
This update in the model allows to modulate its behavior according to the current regime, providing increased flexibility.
The introduction of this regime switching mechanism in the GRU architecture is  suitable for tasks where the dataset is subject to context variations, which is often the case in time series and particularly in highly volatile assets. By adapting its dynamics to the current regime, the Switching GRU can better handle the complex underlying patterns present in such data. The succession of operations detailed in \eqref{eq:k_gru} are used to estimate log returns of each possible states. On the right side of the Figure \ref{m-GRU} we have another GRU. We will denote $\text{GRU}_{k,in}$ the successive equations \eqref{eq:k_gru}, the other GRU (on the right side) will be denoted $GRU_c$ which will be fed using the covariate only and not the entire datasets containing covariates and past observed values of the target. The $\text{GRU}_c$ will transform covariates to $Z_t$ vector such
\begin{equation}
    Z_t = \text{GRU}_c (\Fc_{Z,t-1})
\end{equation}
The $Z_t$ and the vector of $\Vec{\hat{r_t}}=(r_{1,t},r_{2,t},...,r_{m,t})$ obtained previously with the use of $\text{GRU}_{k,in}$ will be the input of our switching mechanism detailed \ref{sec:switching_mechanism} to estimate the vector $\Vec{\pi_t}=(\pi_{1,t},\pi_{2,t},...,\pi_{m,t})$. The ouput of framework is $\Vec{\pi_t}$. In our regime-switching model, the function $S(r_t, Z_t)$ generates a vector of probabilities $\pi_t$ for $m$ regimes at time $t$. For a two-regime system, $\pi_t = (\pi_{t,1}, \pi_{t,2})$. The predicted regime $\hat{y}_t$ is determined by $\hat{s}_t = \argmax_k(\pi_{t,k})$, which can be expressed by an indicator function $I_{t,k}$. The true regime $s_t \in \{0,1\}$ is one-hot encoded. The confusion matrix $C$ is then constructed as $C_{ij} = \sum_t I(\hat{s}_t = i \text{ and } s_t = j)$, where $\hat{s}_t$ is the predicted regime and $s_t$ what we defined as a true regime.
\begin{figure}[H]
\centering
\includegraphics[width=12.0cm]{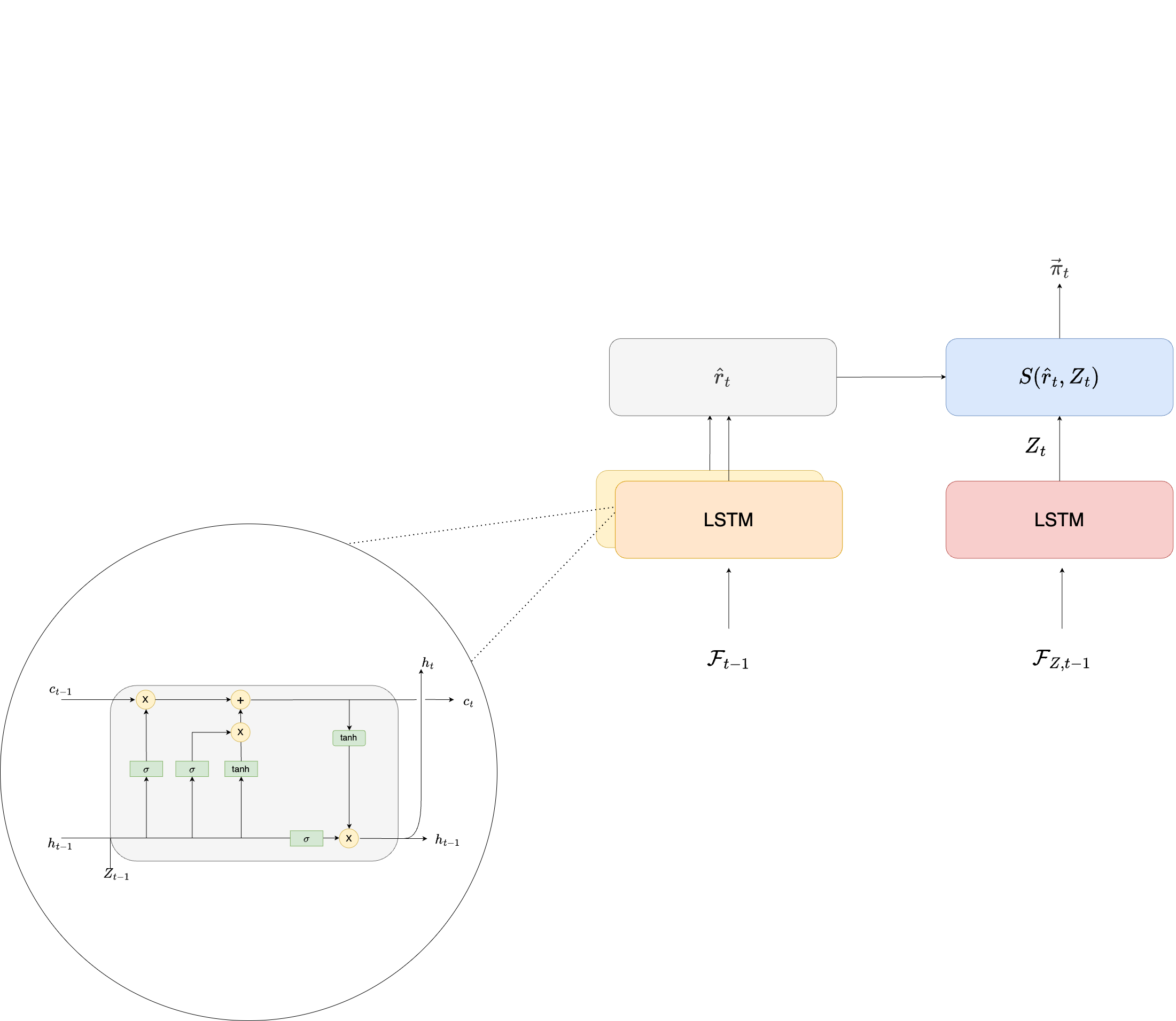}
\caption{Structure of the m-LSTM}
\label{m-lstm}
\end{figure}

\medskip

The methodology would be the same for the LSTM \cite{lstm}. m-LSTM are stacked with the following ouputs $ \{h_{k,t}\}_{k=1}^{m} := \{h_{1,t},h_{2,t},...,h_{m,t}\}$ and a set of parameters $ \{\theta_k\}_{k=1}^{m} :=  \{\theta_1,\theta_2,...,\theta_m\}$ where; $$\theta_k := \{W_{k,f},W_{k,i},W_{k,c},W_{k,o},b_{k,f},b_{k,i},b_{k,c},b_{k,o}\},$$Our model, described in Figure \ref{m-lstm} shows the LSTM stacked on the left side. We introduce the notation $\text{LSTM}_{k,in}$ to denoted the succession of the following operation,
\begin{equation}
\label{eq:k_lstm}
\begin{split}
    f_{k,t} & = \sigma(W_{k,f}\cdot[h_{k,t-1},z_{t-1}]+ b_{k,f}),\\
    i_{k,t}& = \sigma(W_{k,i}\cdot[h_{k,t-1},z_{t-1}]+ b_{k,i}),\\
    {\tilde c}_{k,t} & = {\tanh (W_{k,c}\cdot[h_{k,t-1},z_{t-1}]+ b_{k,c})},\\
    c_{k,t} & = f_{k,t} * c_{k,t-1}+i_{k,t} * {\tilde c}_{k,t},\\
    o_{k,t} & = \sigma(W_{k,o}\cdot[h_{k,t-1},z_{t-1}]+ b_{k,o}),\\ 
    h_{k,t} & = o_{k,t} * {\tanh(c_{k,t})}.
\end{split}
\end{equation}
In the same way as the m-GRU, the m-LSTM has another LSTM, called $LSTM_c$, which will be used to encode the covariates that will be used to estimate $\Vec{\pi_t}$. The third and final model proposed for our framework is the Temporal Kolmogorov Arnold Networks (TKAN),
The TKAN have been introduced in a previous paper \cite{genet2024tkan}. Its excellent results shown in temporal transformer architecture \cite{genet2024temporal} have motivated us to propose a switching extension based on this new architecture. As we have done previously, we will use the notation m-TKAN to designate switching TKAN. Keeping the same notation as the one used for the GRU and the LSTM, $\text{TKAN}_{k,in}$ would be given by,
\begin{equation}
    \begin{split}
      f_{k,t} &= \sigma(W_{k,f} x_t + U_{k,f} h_{k,t-1} + b_{k,f}),\\
      i_{k,t} &= \sigma(W_{k,i} x_t + U_{ki} h_{t-1} + b_{k,i}),\\
        r_{k,t} &= \text{Concat}[\phi_{k,1}(s_{k,1,t}),\phi_{k,2}(s_{k,2,t}),...,\phi_{k,L}(s_{k,L,t})],\\
        o_{k,t} &= \sigma(W_{k,o}r_{k,t} + b_{k,o}),\\
        c_{k,t} &= f_{k,t} \odot c_{k,t-1} + i_{k,t} \odot \tilde{c}_{k,t},\\
        h_{k,t} &= o_{k,t} \odot \tanh(c_{k,t}),
    \end{split}
\end{equation}
where $s_{k,l,t}=W_{k,l,\tilde{x}} x_t + W_{k,l,\tilde{h}} \tilde{h}_{k,l,t-1}$ is the input of each RKAN, $\tilde{c}_{k,t} = \sigma(W_c x_t + U_c h_{t-1} + b_c)$ represents its internal memory, and $\phi_{k,l}$ is a KAN layer of regime $k$. The "memory" step \(\tilde{h}_{k,l,t}\) is defined as a combination of past hidden states for each $k$ regime, such,
\begin{equation}
\tilde{h}_{k,l,t} =  W_{hh} \tilde{h}_{k,l,t-1} + W_{k,hz} \tilde{o}_{t},
\label{eq:update_state_tkan}
\end{equation}
Similar to the m-GRU and m-LSTM, the m-TKAN model has a TKAN layer called $TKAN_c$. This layer is responsible for encoding the covariates, an additional input variables denoted $2_t$, to estimate the probabilities $\Vec{\pi_t}$. After estimating the probabilities and deducing the predicted regimes, we backtest a simple strategy for each of the models. Depending on the prediction made for $\hat{s}_t$, we open a long or short position if the predicted regime is bullish or bearish. All the results of these backtests are available in Appendix \ref{subsec:switching_gru}-\ref{subsec:switching_lstm}-\ref{subsec:switching_tkan}.
\begin{figure}[H]
\centering
\includegraphics[width=12.0cm]{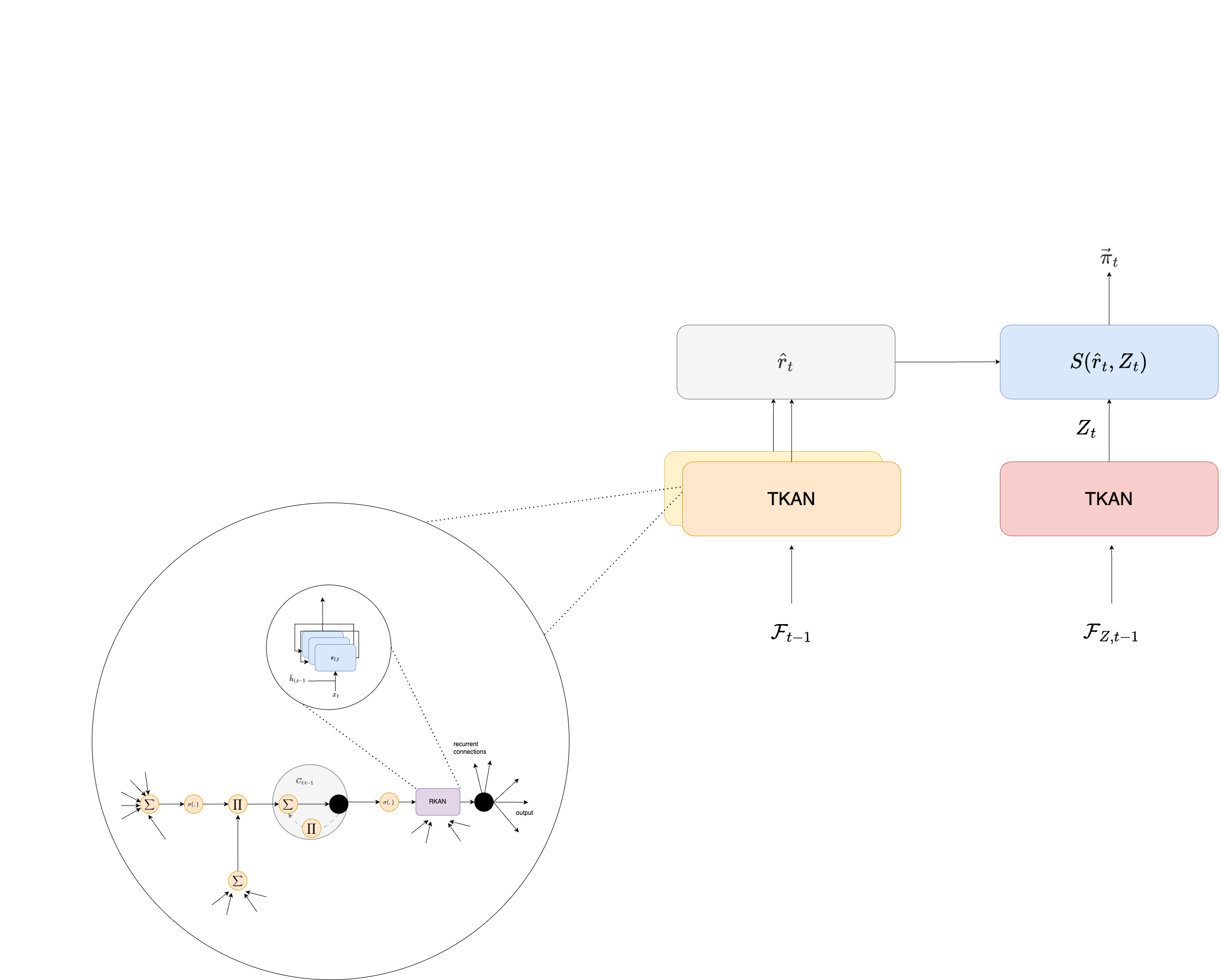}
\caption{Structure of the m-TKAN}
\label{m-lstm}
\end{figure}
\textit{The following section section will also examines the results obtains for these backtests.}

\subsection{Learning Task}
To proceed  the estimation of regime switching parameters, we download the open, high, low, close (OHLC) of Bitcoin from \href{https://www.cryptocompare.com}{cryptocompare.com} and we compute the log return defined as:
$$
r_t = log(p_t) - log(p_{t-1}),
$$
then we build our HML and IV \eqref{eq:iv}. These coviariates will be stacked in a vector with a fixed sequence length. The input vector $Z_{0:T-1}$ will be standardized and divided by the maximum of absolute value.
$$Z_{0:T-1}:=\{\{HML_t\}_{t=0}^{T-1},\{IV_t\}_{t=0}^{T-1}\}.$$ 
Creating a learning task to predict regime is not as straightforward as it is for many other models, as the real states are not known. In order to test wether our model is efficient in predicting, we had to labelize our datas with regime. To do so, we use a simple systematic methods that consists as defining two regimes, a bull one when the average price of the 20 past days (including the observation at t when we make our prediction) is lower than the average price of the 20 next days, a bear one in the opposite case. Having such, the tasks become a standard classifcation tasks, where we have to predict in which class we are, given the current information in $t$.
\medskip
The outputs of the model being two probability, we encode the classes using a one hot encoder, and thus calibrate the model using a categorical cross-entropy as loss, which is the standard for this kind of problem. The Categorical cross-entropy
$$
L = -\sum_{i=1}^{N} \sum_{c=1}^{C} y_{ic} \log(p_{ic}) 
$$
where $N$ is the number of samples, $y_{ic}$ is the binary indicator (0 or 1) indicating whether the class label $c$ is the correct classification for sample $i$, and $p_{ic}$ is the predicted probability that the sample $i$ belongs to the class $c$. For each sample $i$ and each class denoted $c$, the binary indicator $y_{ic}$ is 1 if the sample belongs to class $c$ and 0 otherwise. This loss function commonly used in classification tasks to measure the difference between two probability distributions: true labels and predicted labels. Finally, we used a validation set during the training, in order to use an early stopping callback that stop the training after 10 consecutive epochs without improvement of the validation loss, as well as a learning rate reduction by a factor of 4 after 5 consecutive epochs without improvments. These two together reduce the risks of overfitting and enables to have a systematic approach on this learning rate selection. We used RNNs as the neural network parts in our model, as we added a sequence dimension to the input, in order to be able to represent the markov-chain. We thus compared the two most standards RNN that are the GRU and LSTM, but also the TKAN. Finally, in order to test the different RNNs the same way, we build all the model the same, with 2 layers of 100 units in each, using their standard activation functions. Only the TKAN as a bit more hyper-parameter, with 3 internal RKAN layers of degree 3 and grid-size 5 which are the defaults of the models, and an internal KAN sub-layers output dimension of 10.

\medskip

During the training task, we know that neural networks can tend to adapt to the training data and this is due to the large number of iterations and then become unable to generalise what they have learned on the training phase to perform well on the test set.
One way to overcome this problem is to track the evolution of the error on the training and validation sets at each iteration and analytically find out at which iteration the error on the test set increases while that on the training set continues to decrease.
This technique allows us to select the parameters of our model without overfitting bias.

\medskip

We do not seek to fine-tune our extended models but to assess the ability of our predictor to identify market regimes, and predict the next market regime. We also seek  to stabilise the transition probabilities which seems to be very sensitive to the presence of some covariates during the estimation process for the switching markov models. We believe that neural networks will help stabilize the probability of avoiding the transition from one state to another. Indeed, the complexity and the number of coefficients that we will have in the most complete models will reduce this increased sensitivity to covariates. For the learning task we have built a training set of 80\% of the total observations we have and among this 80\% we use 20\% of this training set to build a validation with an early stopping mechanism in order to avoid overfitting.

\section{Results}

Looking at the results on test set, it appears that the m-GRU model has a high number of false positives compared to true negatives. This indicates it tends to incorrectly classify negatives as positives. However, the true positives are higher than false negatives. This suggest it performs better in correctly identifying positive cases. The m-LSTM model shows a better performance with fewer false positives and a higher number of true negatives compared to the GRU model. However, it has a slightly higher number of false negatives and fewer true positives. It indicates a bit of a trade-off in correctly identifying positive cases. The TKAN model shows the best performance in terms of minimizing false positives. TKAN obtained the highest number of true negatives among the three models. It also maintains a reasonable balance between false negatives and true positives, indicating a good overall performance in identifying both positive and negative cases accurately. Results obtained during the training task are available in the appendix.

\begin{figure}[H]
\minipage{0.32\textwidth}
  \includegraphics[width=\linewidth]{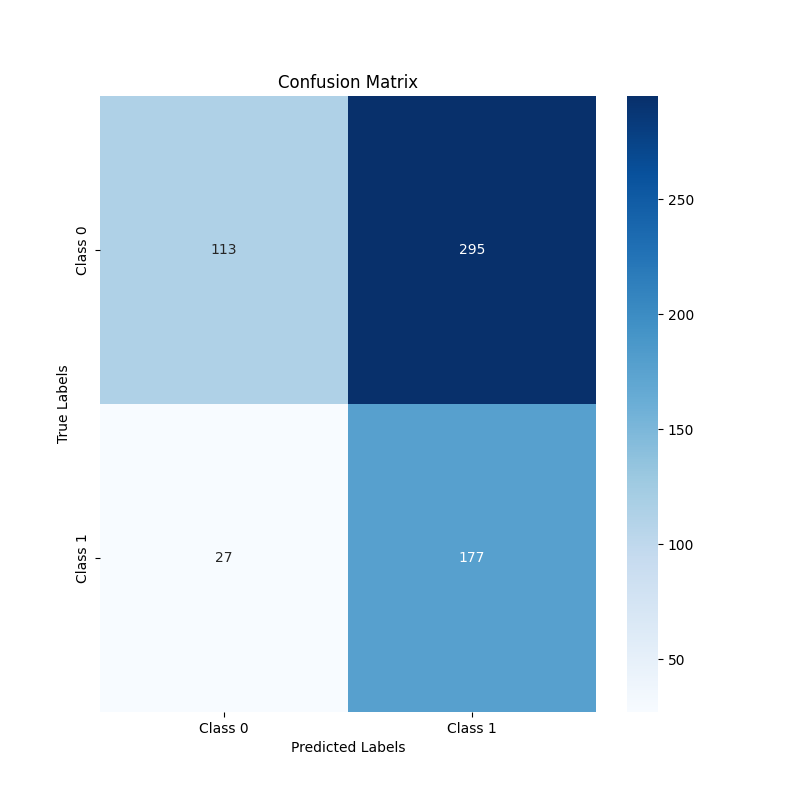}
  \caption{GRU Confusion matrix (out of sample)}\label{fig:awesome_image1}
\endminipage\hfill
\minipage{0.32\textwidth}
  \includegraphics[width=\linewidth]{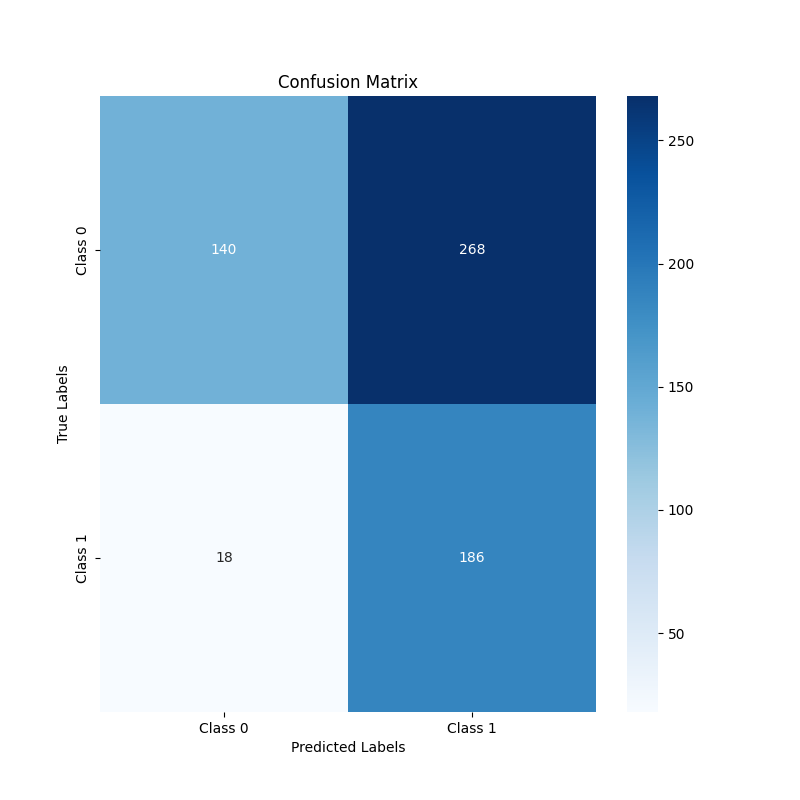}
  \caption{LSTM Confusion matrix (out of sample)}\label{fig:awesome_image2}
\endminipage\hfill
\minipage{0.32\textwidth}%
  \includegraphics[width=\linewidth]{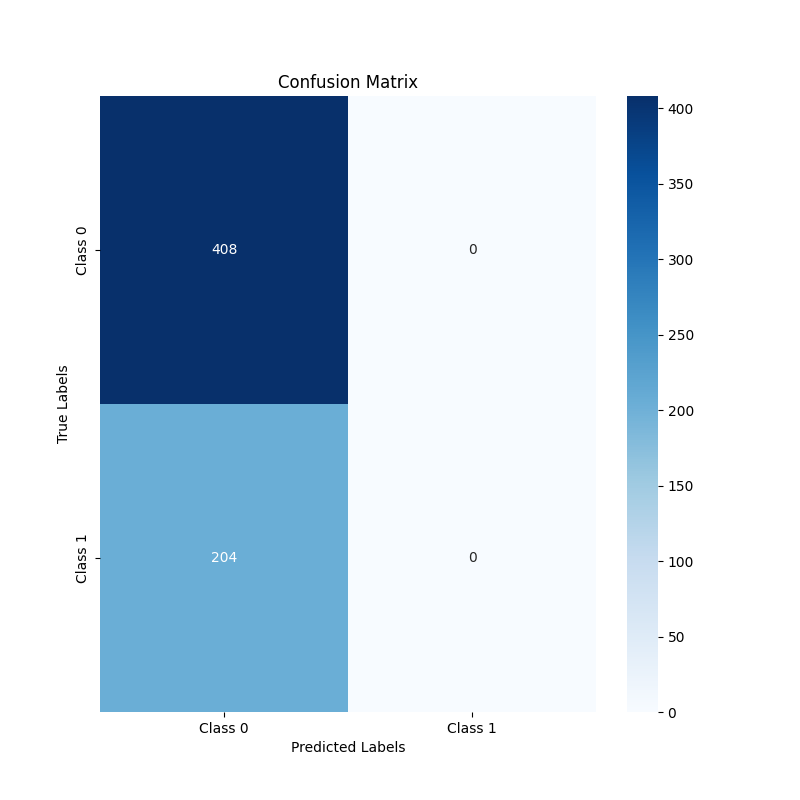}
  \caption{TKAN Confusion matrix (out of sample)}\label{fig:awesome_image3}
\endminipage
\end{figure}

\begin{figure}[H]
\minipage{0.32\textwidth}
  \includegraphics[width=\linewidth]{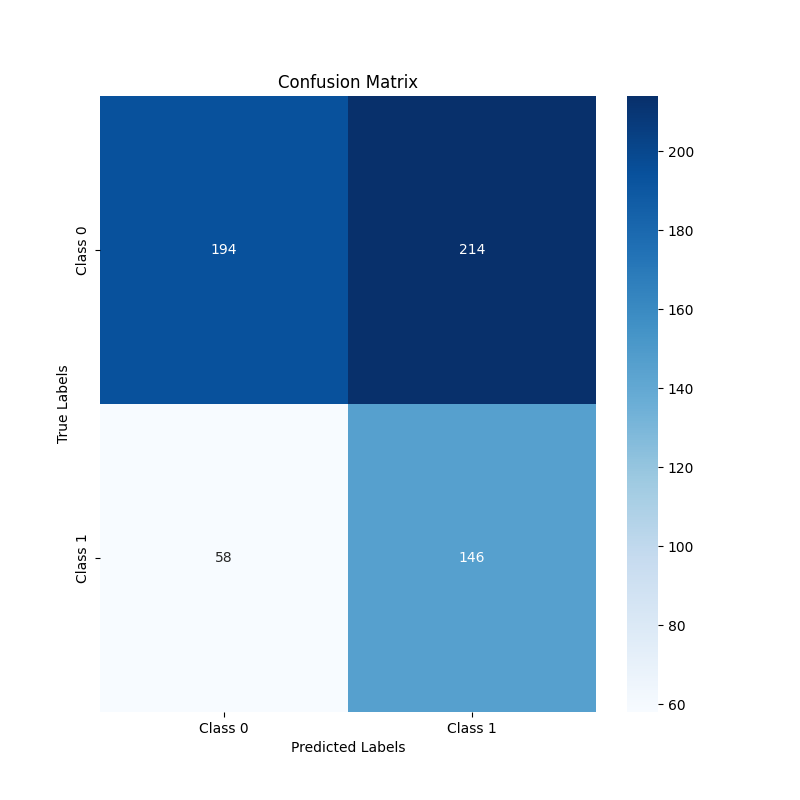}
  \caption{m-GRU Confusion matrix (out of sample)}\label{fig:awesome_image1}
\endminipage\hfill
\minipage{0.32\textwidth}
  \includegraphics[width=\linewidth]{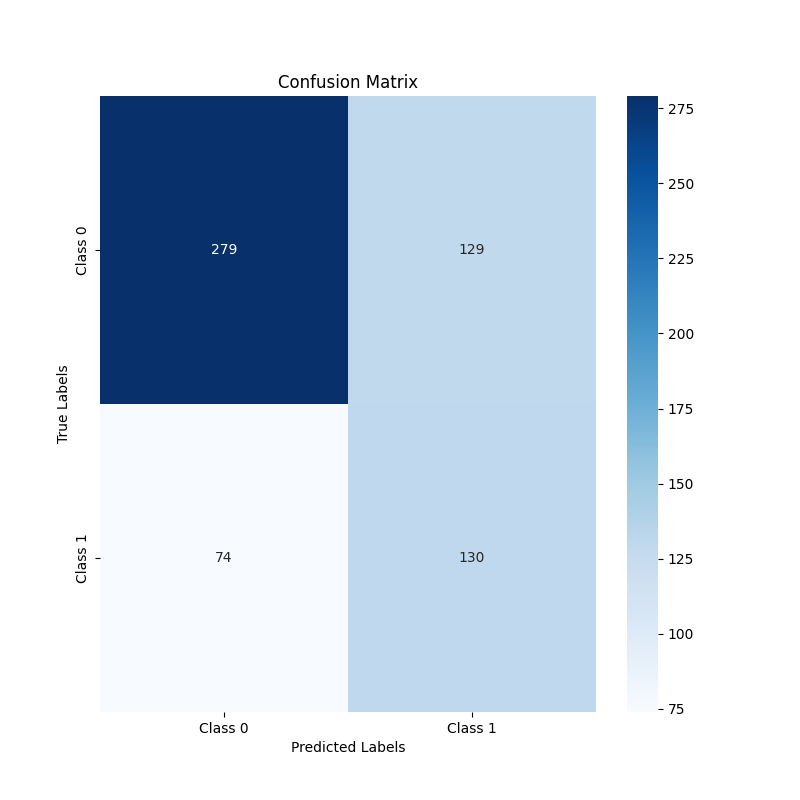}
  \caption{m-LSTM Confusion matrix (out of sample)}\label{fig:awesome_image2}
\endminipage\hfill
\minipage{0.32\textwidth}%
  \includegraphics[width=\linewidth]{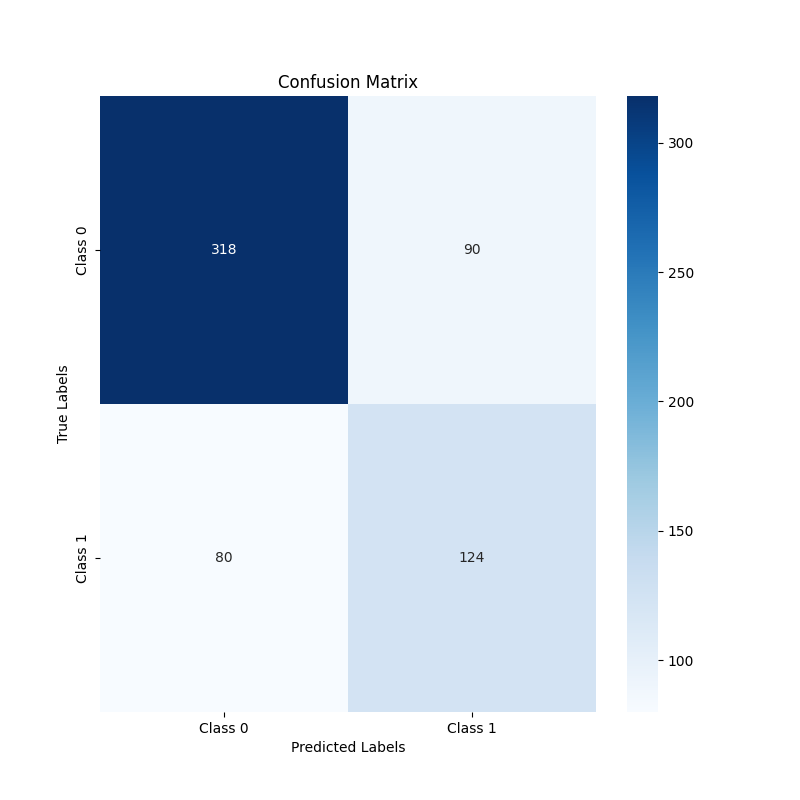}
  \caption{m-TKAN Confusion matrix (out of sample)}\label{fig:awesome_image3}
\endminipage
\end{figure}

\begin{table}[H]
\centering
\caption{Comparison of simple RNNs versus Switching RNNs}
\begin{tabular}{@{}llccccccc@{}}
\toprule
\textbf{Model} & \textbf{Class} & \textbf{Precision} & \textbf{Recall} & \textbf{F1-Score} & \textbf{Support} & \textbf{Accuracy} \\
\midrule
\multicolumn{7}{c}{LSTM vs. m-LSTM} \\
\midrule
LSTM    & Class 0 & \textbf{0.89} & 0.34 & 0.49 & 408 & \multirow{2}{*}{0.53} \\
        & Class 1 & 0.41 & \textbf{0.91} & 0.57 & 204 &  \\
m-LSTM  & Class 0 & 0.79 & \textbf{0.68} & \textbf{0.73} & 408 & \multirow{2}{*}{\textbf{0.67}} \\
        & Class 1 & \textbf{0.50} & 0.64 & \textbf{0.56} & 204 &  \\
\midrule
\multicolumn{7}{c}{GRU vs. m-GRU} \\
\midrule
GRU     & Class 0 & \textbf{0.81} & 0.28 & 0.41 & 408 & \multirow{2}{*}{0.47} \\
        & Class 1 & 0.38 & 0.87 & 0.52 & 204 &  \\
m-GRU   & Class 0 & 0.77 & \textbf{0.48} & \textbf{0.59} & 408 & \multirow{2}{*}{\textbf{0.56}} \\
        & Class 1 & \textbf{0.41} & \textbf{0.72} & \textbf{0.52} & 204 &  \\
\midrule
\multicolumn{7}{c}{TKAN vs. m-TKAN} \\
\midrule
TKAN    & Class 0 & 0.67 & \textbf{1.00} & \textbf{0.80} & 408 & \multirow{2}{*}{0.67} \\
        & Class 1 & 0.00 & 0.00 & 0.00 & 204 &  \\
m-TKAN  & Class 0 & \textbf{0.79} & 0.78 & 0.79 & 408 & \multirow{2}{*}{\textbf{0.72}} \\
        & Class 1 & \textbf{0.58} & \textbf{0.59} & \textbf{0.58} & 204 &  \\
\bottomrule
\end{tabular}
\label{tab:comp_accuracy}
\end{table}

Table \ref{tab:comp_accuracy} shows that our Switching neural networks enhance the model's capacity to learn and predict meaningful regimes. Indeed, conventional RNNs are unable to idenitfy market regime, as evidenced by the superior performance of TKAN, which merely identifies the dominant class. However, our Switching models demonstrate higher performance in forecasting for all models. Despite the quality of GRU-based models, those incorporating TKAN units exhibit notable accuracy. The TKAN-based model, in particular, demonstrates robust performance on external tasks.

\begin{table}[H]
    \centering
    \begin{tabular}{l|l|l|l}
         & m-GRU & m-LSTM & m-TKAN  \\ \hline
        Mean Return ($\mu$) & 0.754659 & 0.794480 & \textbf{1.131593}  \\ 
        Standard Deviation ($\sigma$) & 0.734199 & 0.734084 & 0.732871  \\ 
        Sharpe Ratio & 1.027867 & 1.082273 & \textbf{1.544053}  \\ 
        Max Drawdown & -1.071958 & -1.254106 & -0.740906  \\ 
        Sortino Ratio & 1.783035 & 1.863751 & 2.690941  \\ 
        Mean Daily Turnover & 0.173849 & 0.109761 & 0.360958  \\ 
        Annual Turnover & 63.454880 & 40.062615 & 131.749540  \\ 
        Mean Return on Volume & 0.011893 & 0.019831 & 0.008589  \\ 
        Beta & -0.142187 & -0.045676 & 0.154983  \\ 
        Alpha & 0.880322 & 0.834848 & 0.994620 \\ 
        \hline
    \end{tabular}
        \caption{Performance table (In Sample)}
    \label{tab:train_switching}
\end{table}

Table \ref{tab:train_switching} shows the results obtained on the training task (in sample estimation) using the m-GRU, m-LSTM and m-TKAN. Results show significant big differences in their predictive capabilities between train and test set. During the training phase, TKAN stands out for its high average return (1.131593) and a higher Sharpe (1.544053) and Sortino (2.690941) ratios. These ratios suggest a significantly better risk-adjusted performance than the other models. The m-LSTM, demonstrating a Sharpe ratio of (1.082273) and a Sortino ratio of (1.863751), also performed commendably, although it remains slightly inferior to the m-TKAN. Conversely, the m-GRU encountered more challenges in its performance. Despite a good average return (0.754659), it has a lower Sharpe ratio (1.027867) and Sortino ratio (1.783035), as well as a higher MDD (-1.071958).

\begin{table}[H]
    \centering
    \begin{tabular}{l|l|l|l}
         & m-GRU & m-LSTM & m-TKAN  \\ \hline
        Mean Return ($\mu$) & -0.398766 & 0.198593 & \textbf{0.444902}  \\ 
        Standard Deviation ($\sigma$) & 0.490487 & 0.490821 & 0.490378  \\ 
        Sharpe Ratio & -0.813001 & 0.404614 & \textbf{0.907263}  \\ 
        Max Drawdown & -0.687949 & -0.342666 & -0.467687  \\ 
        Sortino Ratio & -1.219218 & 0.659887 & 1.497058  \\ 
        Mean Daily Turnover & 0.173486 & 0.163666 & 0.468085  \\ 
        Annual Turnover & 63.322422 & 59.738134 & 170.851064  \\ 
        Mean Return on Volume & -0.006297 & 0.003324 & 0.002604  \\ 
        Beta & -0.043611 & 0.156359 & 0.374576  \\ 
        Alpha & -0.361273 & 0.064168 & 0.122870 \\
        \hline
    \end{tabular}
    \caption{Performance table (Out of Sample)}
    \label{tab:test_switching}
\end{table}
The Table \ref{tab:test_switching} reveals impressive metrics for the m-TKAN during testing (out of sample estimation). The m-LSTM also does very well on the test sample. The TKAN maintained good risk control, with a moderate max drawdown (-0.467687) and a high Sortino ratio (1.497058). The m-GRU, on the other hand, shows a negative test performance with an average return of -0.398766 and unfavorable risk ratios, underlining a poorer ability to generalize. The m-LSTM and m-TKAN appear to be more robust and efficient models for managing sequential data in a variety of market environment, with m-TKAN standing out in particular for its ability to maximize risk-adjusted returns.

\section{Conclusion}
In conclusion, we have seen that switching models are particularly beneficial for analyzing digital assets. Their relevance can be explained by the highly volatile dynamic nature of this new market, still in its beginning. These models are very effective at capturing rapid transitions between different states (bullish or bearish). They adapt quickly and efficiently to the influence of external factors such as regulatory or technological changes. They are able to track the structural evolution of this market. These models provide a robust analytical framework for understanding the complex dynamics of the digital asset market. In this paper, we proposed the incorporation of Markov switching into recurrent neural network models, an innovative framework that improves the performance of these models. Particularly in the case of TKAN, this new state-space framework allows for a substantial improvement in the ability to capture and predict significant regimes in sequential data. The m-TKAN shows the most significant improvement, evolving from a model unable to classify class 1 to a model performing well on both classes. In the context of financial data, this improvement translates into superior financial performance and better risk management. Looking at the other models, those incorporating the markov switching framework (m-LSTM, m-GRU, m-TKAN) demonstrate better overall predictive capacity than their conventional counterparts. The Markov switching models tend to offer more balanced performance between classes, whereas the classical models tend to favor one class over another. This study underlines the effectiveness of integrating Markov chain structures into recurrent neural network models, enhancing their ability to process complex sequential data and identify different regimes or classes.

\bibliography{bib.bib}
\section{Appendix}

\subsection{Regime Switching without TVTP (3 states)}
\label{regime_3}

\begin{figure}[H]
   \centering
    \includegraphics[scale=0.4]{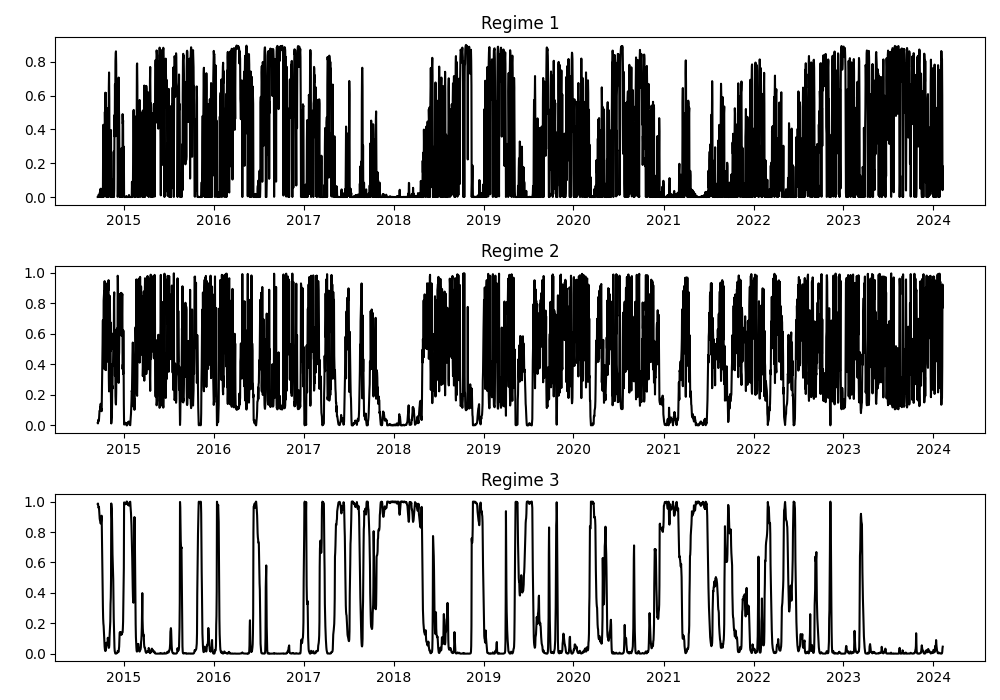}
    \caption{Smoothed Marginal Probabilities}
    \label{fig:3_regimes_model_without_tvtp}
\end{figure}

\begin{center}

\begin{tabular}{lcccccc}
\toprule
                & \textbf{coef} & \textbf{std err} & \textbf{z} & \textbf{P$> |$z$|$} & \textbf{[0.025} & \textbf{0.975]}  \\
\midrule
\textbf{const}  &       0.0007  &        0.000     &     1.544  &         0.123        &       -0.000    &        0.002     \\
\textbf{sigma2} &    7.957e-05  &     2.22e-05     &     3.585  &         0.000        &     3.61e-05    &        0.000     \\
                & \textbf{coef} & \textbf{std err} & \textbf{z} & \textbf{P$> |$z$|$} & \textbf{[0.025} & \textbf{0.975]}  \\
\midrule
\textbf{const}  &       0.0028  &        0.001     &     2.433  &         0.015        &        0.001    &        0.005     \\
\textbf{sigma2} &       0.0010  &        0.000     &     4.466  &         0.000        &        0.001    &        0.001     \\
                & \textbf{coef} & \textbf{std err} & \textbf{z} & \textbf{P$> |$z$|$} & \textbf{[0.025} & \textbf{0.975]}  \\
\midrule
\textbf{const}  &       0.0003  &        0.002     &     0.137  &         0.891        &       -0.004    &        0.005     \\
\textbf{sigma2} &       0.0030  &        0.000     &     9.709  &         0.000        &        0.002    &        0.004     \\
                   & \textbf{coef} & \textbf{std err} & \textbf{z} & \textbf{P$> |$z$|$} & \textbf{[0.025} & \textbf{0.975]}  \\
\midrule
\textbf{p[1-$>$1]} &       0.6461  &        0.101     &     6.386  &         0.000        &        0.448    &        0.844     \\
\textbf{p[2-$>$1]} &       0.2840  &        0.116     &     2.442  &         0.015        &        0.056    &        0.512     \\
\textbf{p[3-$>$1]} &     4.88e-06  &        0.112     &  4.37e-05  &         1.000        &       -0.219    &        0.219     \\
\textbf{p[1-$>$2]} &       0.3306  &        0.074     &     4.440  &         0.000        &        0.185    &        0.477     \\
\textbf{p[2-$>$2]} &       0.6970  &        0.104     &     6.721  &         0.000        &        0.494    &        0.900     \\
\textbf{p[3-$>$2]} &       0.0524  &        0.080     &     0.657  &         0.511        &       -0.104    &        0.209     \\
\bottomrule
\end{tabular}
\end{center}

\newpage

\subsection{Regime Switching TVTP (3  states)}
\label{regimeTVTP_3}

\subsubsection*{Regime Switching HML, TVTP with HML Factor}

\begin{center}

\begin{table}[H]
    \centering
    \begin{tabular}{lcccccc}
\toprule
                & \textbf{coef} & \textbf{std err} & \textbf{z} & \textbf{P$> |$z$|$} & \textbf{[0.025} & \textbf{0.975]}  \\
\midrule
\textbf{const}  &       0.0003  &        0.002     &     0.165  &         0.869        &       -0.003    &        0.004     \\
\textbf{hml}     &      -0.0010  &        0.003     &    -0.355  &         0.723        &       -0.007    &        0.005     \\
\textbf{sigma2} &    7.925e-05  &     4.09e-05     &     1.936  &         0.053        &    -9.65e-07    &        0.000     \\
                & \textbf{coef} & \textbf{std err} & \textbf{z} & \textbf{P$> |$z$|$} & \textbf{[0.025} & \textbf{0.975]}  \\
\midrule
\textbf{const}  &       0.0032  &        0.001     &     2.671  &         0.008        &        0.001    &        0.006     \\
\textbf{hml}     &      -0.0002  &        0.004     &    -0.056  &         0.955        &       -0.008    &        0.008     \\
\textbf{sigma2} &       0.0010  &        0.001     &     1.725  &         0.085        &       -0.000    &        0.002     \\
                & \textbf{coef} & \textbf{std err} & \textbf{z} & \textbf{P$> |$z$|$} & \textbf{[0.025} & \textbf{0.975]}  \\
\midrule
\textbf{const}  &      -0.0001  &        0.003     &    -0.041  &         0.967        &       -0.006    &        0.006     \\
\textbf{hml}     &      -0.0001  &        0.003     &    -0.042  &         0.967        &       -0.006    &        0.006     \\
\textbf{sigma2} &       0.0032  &        0.000     &     9.058  &         0.000        &        0.002    &        0.004     \\
                         & \textbf{coef} & \textbf{std err} & \textbf{z} & \textbf{P$> |$z$|$} & \textbf{[0.025} & \textbf{0.975]}  \\
\midrule
\textbf{p[1-$>$1].const} &       2.1642  &       10.353     &     0.209  &         0.834        &      -18.127    &       22.455     \\
\textbf{p[2-$>$1].const} &       2.5755  &        5.538     &     0.465  &         0.642        &       -8.279    &       13.431     \\
\textbf{p[3-$>$1].const} &      -7.2483  &        3.618     &    -2.004  &         0.045        &      -14.339    &       -0.158     \\
\textbf{p[1-$>$1].hml} &      -2.9672  &        4.644     &    -0.639  &         0.523        &      -12.069    &        6.135     \\
\textbf{p[2-$>$1].hml} &      -0.0084  &        5.774     &    -0.001  &         0.999        &      -11.324    &       11.308     \\
\textbf{p[3-$>$1].hml} &       0.3645  &        1.193     &     0.305  &         0.760        &       -1.974    &        2.703     \\
\textbf{p[1-$>$2].const} &       2.6221  &       10.477     &     0.250  &         0.802        &      -17.913    &       23.157     \\
\textbf{p[2-$>$2].const} &       3.3448  &        5.472     &     0.611  &         0.541        &       -7.381    &       14.070     \\
\textbf{p[3-$>$2].const} &      -2.6787  &        0.581     &    -4.611  &         0.000        &       -3.817    &       -1.540     \\
\textbf{p[1-$>$2].hml} &      -0.6451  &        4.562     &    -0.141  &         0.888        &       -9.587    &        8.297     \\
\textbf{p[2-$>$2].hml} &      -0.0233  &        5.892     &    -0.004  &         0.997        &      -11.572    &       11.526     \\
\textbf{p[3-$>$2].hml} &       0.1108  &        1.370     &     0.081  &         0.936        &       -2.574    &        2.795     \\
\bottomrule
\end{tabular}
\end{table}
\end{center}

\begin{figure}[H]
    \centering
    \includegraphics[scale=0.4]{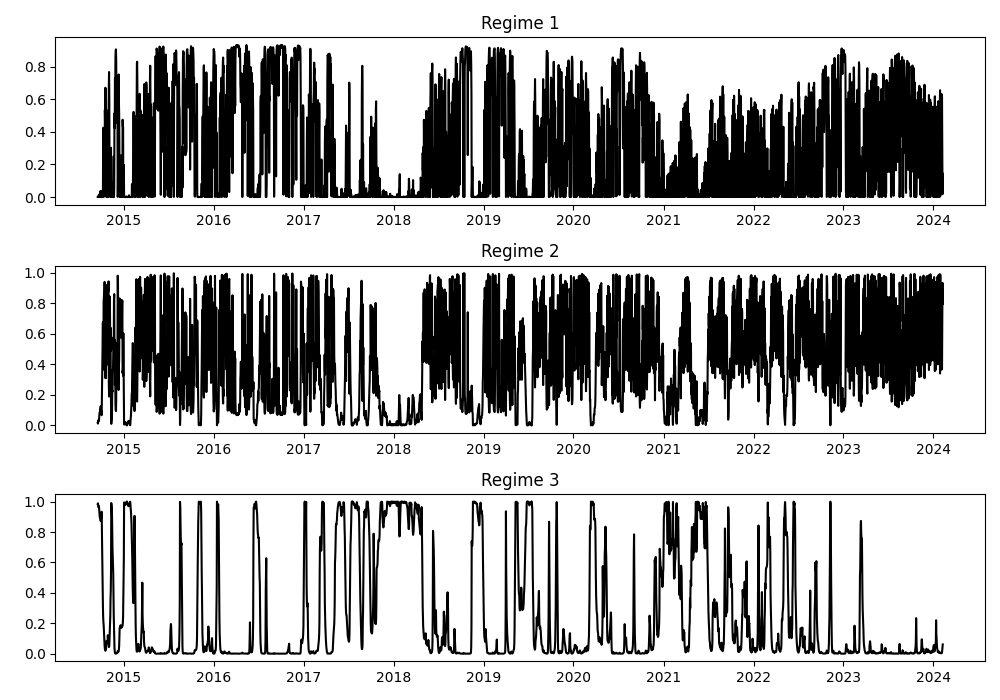}
    \caption{Smoothed Marginal Probabilities}
    \label{fig:3_regimes_model_hml_with_tvtp_hml}
\end{figure}

\subsubsection*{Regime Switching TVTP with HML Factor}

\begin{figure}[H]
    \centering
    \includegraphics[scale=0.4]{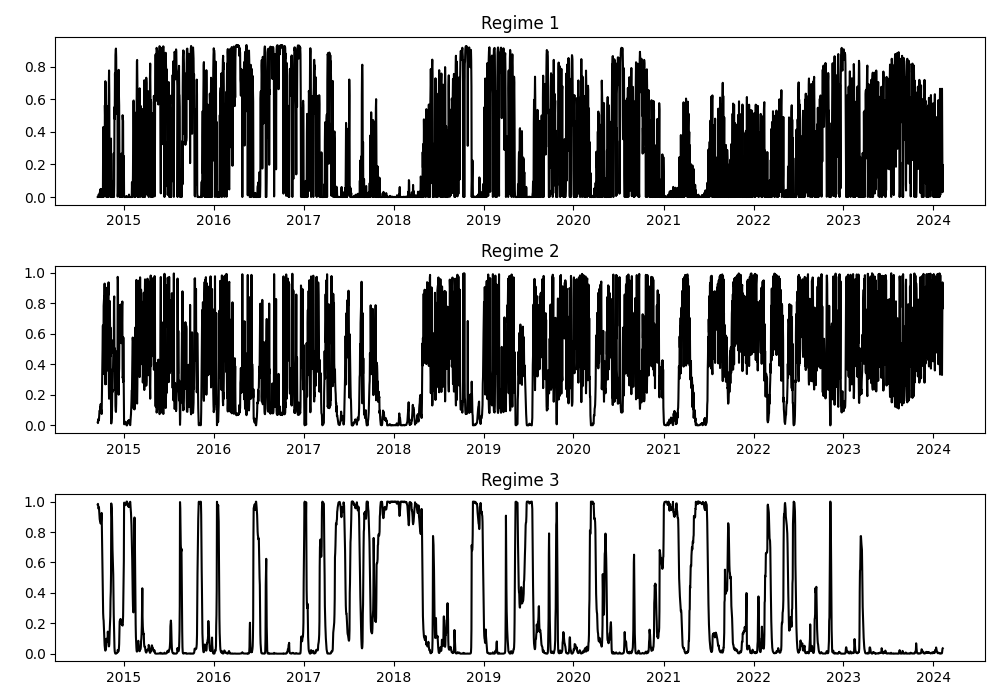}
    \caption{Smoothed Marginal Probabilities}
    \label{fig:3_regimes_model_with_tvtp_hml}
\end{figure}

\begin{center}

\begin{table}[H]
    \centering
    \begin{tabular}{lcccccc}
\toprule
                & \textbf{coef} & \textbf{std err} & \textbf{z} & \textbf{P$> |$z$|$} & \textbf{[0.025} & \textbf{0.975]}  \\
\midrule
\textbf{const}  &       0.0008  &        0.000     &     1.764  &         0.078        &    -8.87e-05    &        0.002     \\
\textbf{sigma2} &    8.456e-05  &     2.23e-05     &     3.784  &         0.000        &     4.08e-05    &        0.000     \\
                & \textbf{coef} & \textbf{std err} & \textbf{z} & \textbf{P$> |$z$|$} & \textbf{[0.025} & \textbf{0.975]}  \\
\midrule
\textbf{const}  &       0.0030  &        0.001     &     2.454  &         0.014        &        0.001    &        0.005     \\
\textbf{sigma2} &       0.0010  &        0.000     &     5.058  &         0.000        &        0.001    &        0.001     \\
                & \textbf{coef} & \textbf{std err} & \textbf{z} & \textbf{P$> |$z$|$} & \textbf{[0.025} & \textbf{0.975]}  \\
\midrule
\textbf{const}  &      -0.0002  &        0.002     &    -0.072  &         0.943        &       -0.005    &        0.005     \\
\textbf{sigma2} &       0.0031  &        0.000     &    10.768  &         0.000        &        0.003    &        0.004     \\
                         & \textbf{coef} & \textbf{std err} & \textbf{z} & \textbf{P$> |$z$|$} & \textbf{[0.025} & \textbf{0.975]}  \\
\midrule
\textbf{p[1-$>$1].const} &       2.3747  &        1.593     &     1.491  &         0.136        &       -0.747    &        5.496     \\
\textbf{p[2-$>$1].const} &       3.3433  &        1.706     &     1.960  &         0.050        &       -0.000    &        6.687     \\
\textbf{p[3-$>$1].const} &      -7.5457  &        4.747     &    -1.590  &         0.112        &      -16.849    &        1.758     \\
\textbf{p[1-$>$1].hml} &      -2.1235  &        2.758     &    -0.770  &         0.441        &       -7.528    &        3.282     \\
\textbf{p[2-$>$1].hml} &       1.0712  &        3.675     &     0.291  &         0.771        &       -6.132    &        8.275     \\
\textbf{p[3-$>$1].hml} &       0.6017  &        0.257     &     2.341  &         0.019        &        0.098    &        1.105     \\
\textbf{p[1-$>$2].const} &       2.6377  &        1.496     &     1.763  &         0.078        &       -0.295    &        5.571     \\
\textbf{p[2-$>$2].const} &       4.0863  &        1.534     &     2.664  &         0.008        &        1.080    &        7.093     \\
\textbf{p[3-$>$2].const} &      -2.8440  &        0.417     &    -6.816  &         0.000        &       -3.662    &       -2.026     \\
\textbf{p[1-$>$2].hml} &      -0.0310  &        2.791     &    -0.011  &         0.991        &       -5.502    &        5.440     \\
\textbf{p[2-$>$2].hml} &       1.1641  &        3.641     &     0.320  &         0.749        &       -5.972    &        8.300     \\
\textbf{p[3-$>$2].hml} &      -0.3378  &        0.369     &    -0.915  &         0.360        &       -1.061    &        0.386     \\
\bottomrule
\end{tabular}
    \caption{Model Parameters}
    \label{tab:my_label}
\end{table}

\end{center}

\subsection{Switching GRU}
\label{subsec:switching_gru}
\begin{figure}[H]
\centering
\includegraphics[width=15.0cm]{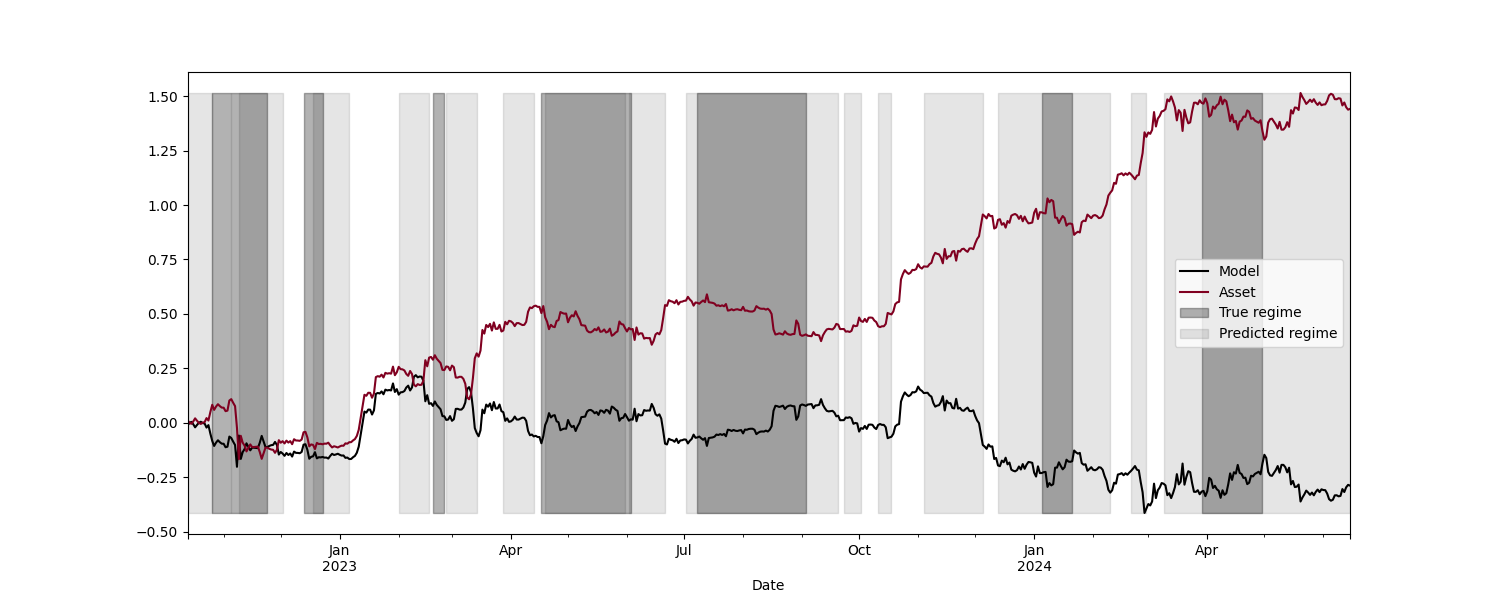}
\caption{GRU (Out of Sample)}
\label{fig:m-gru-train}
\end{figure}

\begin{figure}[H]
\centering
\includegraphics[width=15.0cm]{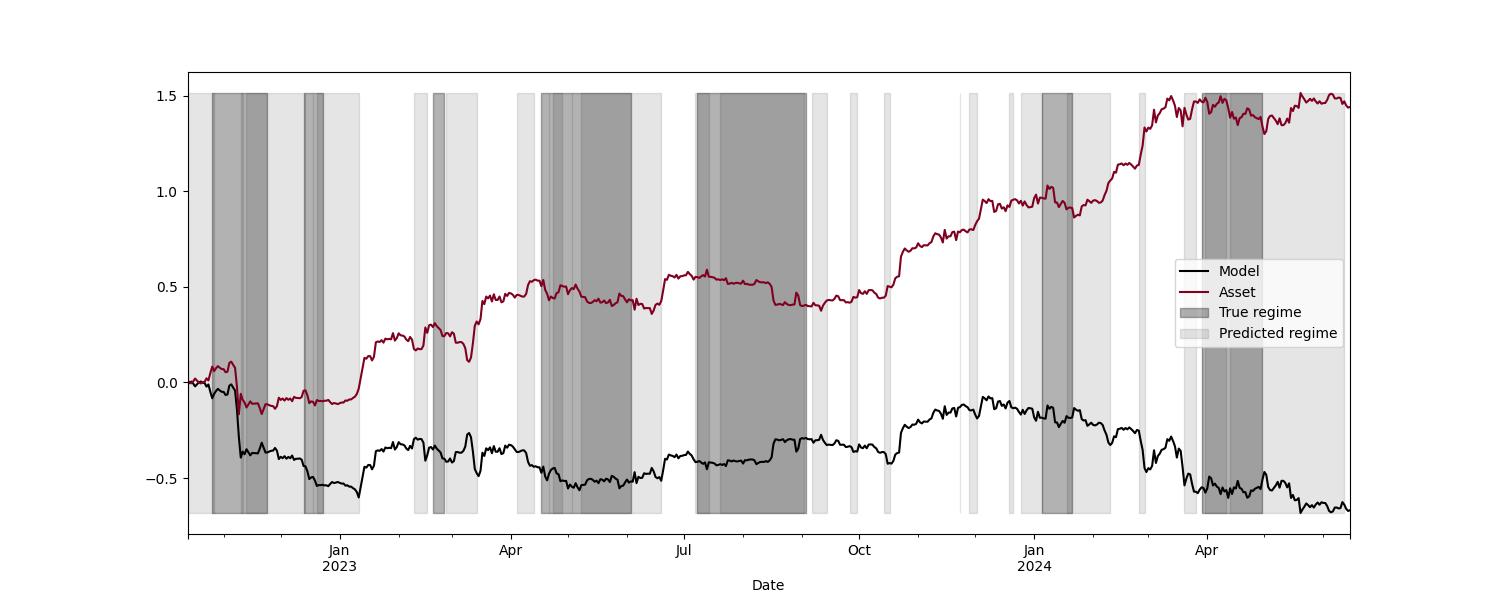}
\caption{m-GRU (Out of sample)}
\label{fig:m-gru-test}
\end{figure}

\subsection{Switching LSTM}
\label{subsec:switching_lstm}

\begin{figure}[H]
\centering
\includegraphics[width=15.0cm]{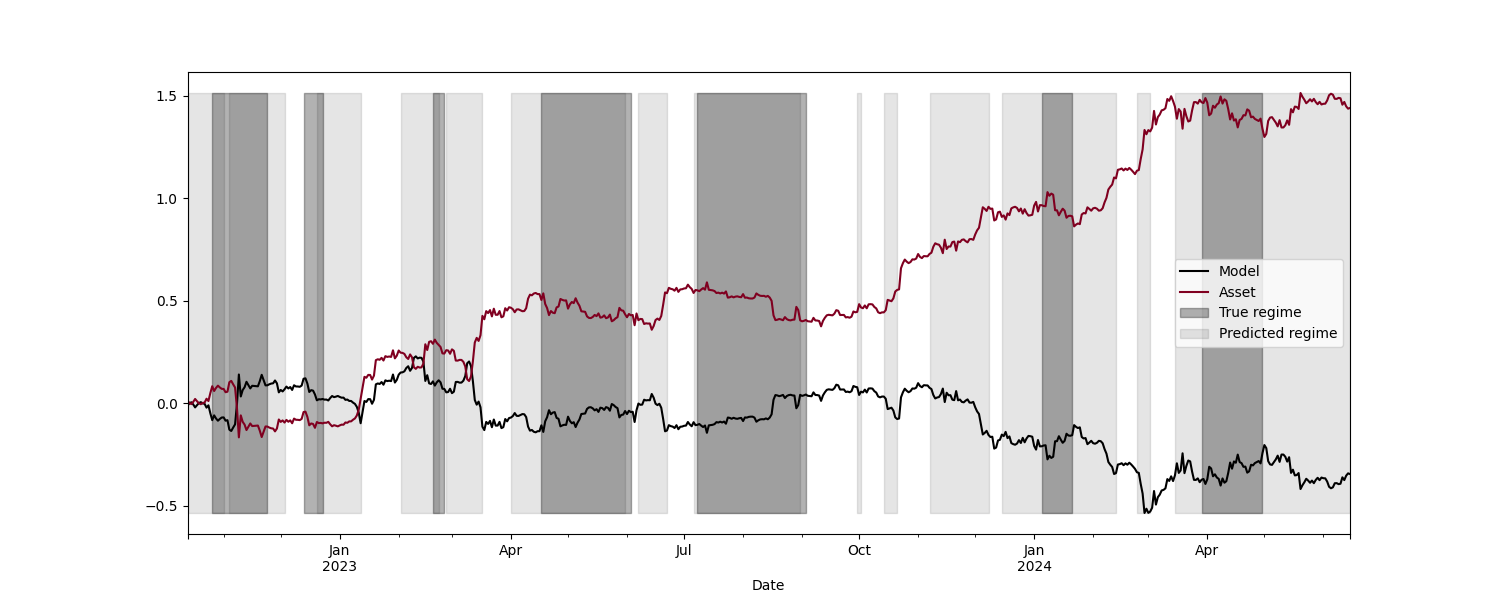}
\caption{LSTM (Out of sample)}
\label{fig:m-lstm-train}
\end{figure}

\begin{figure}[H]
\centering
\includegraphics[width=15.0cm]{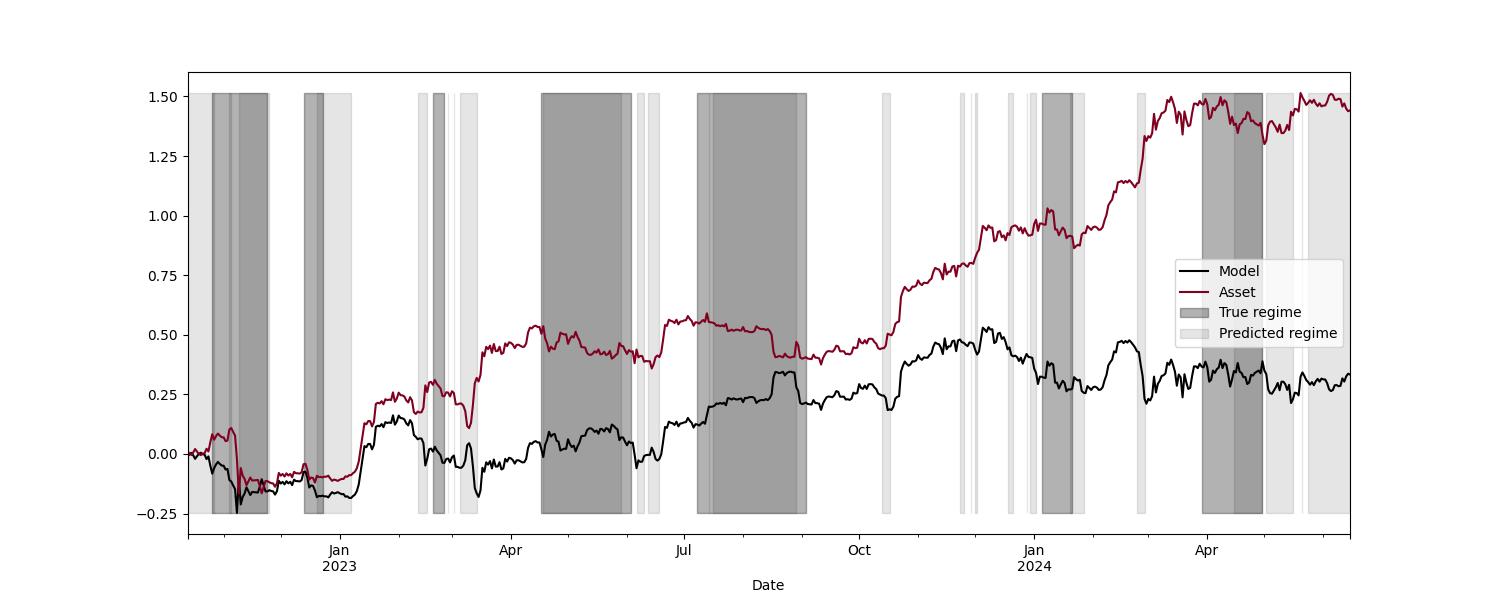}
\caption{m-LSTM (Out of sample)}
\label{fig:m-lstm-test}
\end{figure}

\subsection{Switching TKAN}
\label{subsec:switching_tkan}
\begin{figure}[H]
\centering
\includegraphics[width=15.0cm]{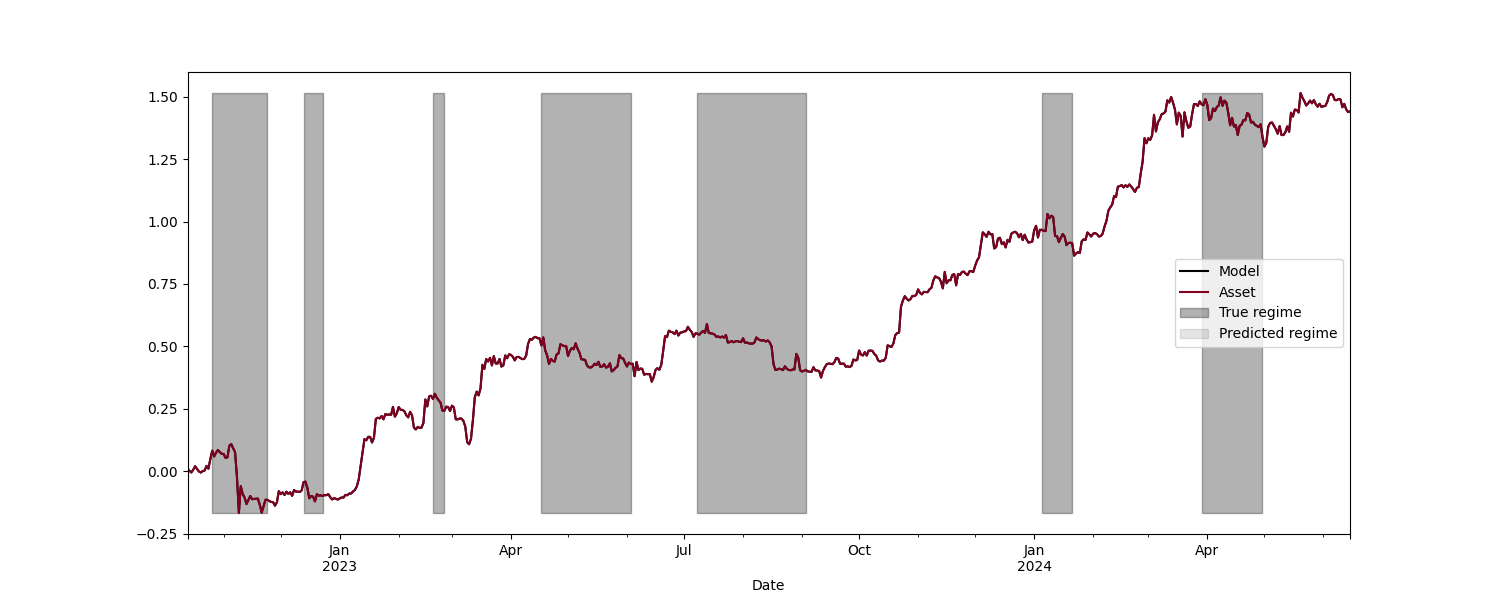}
\caption{TKAN (Out of sample)}
\label{fig:m-tkan-train}
\end{figure}

\begin{figure}[H]
\centering
\includegraphics[width=15.0cm]{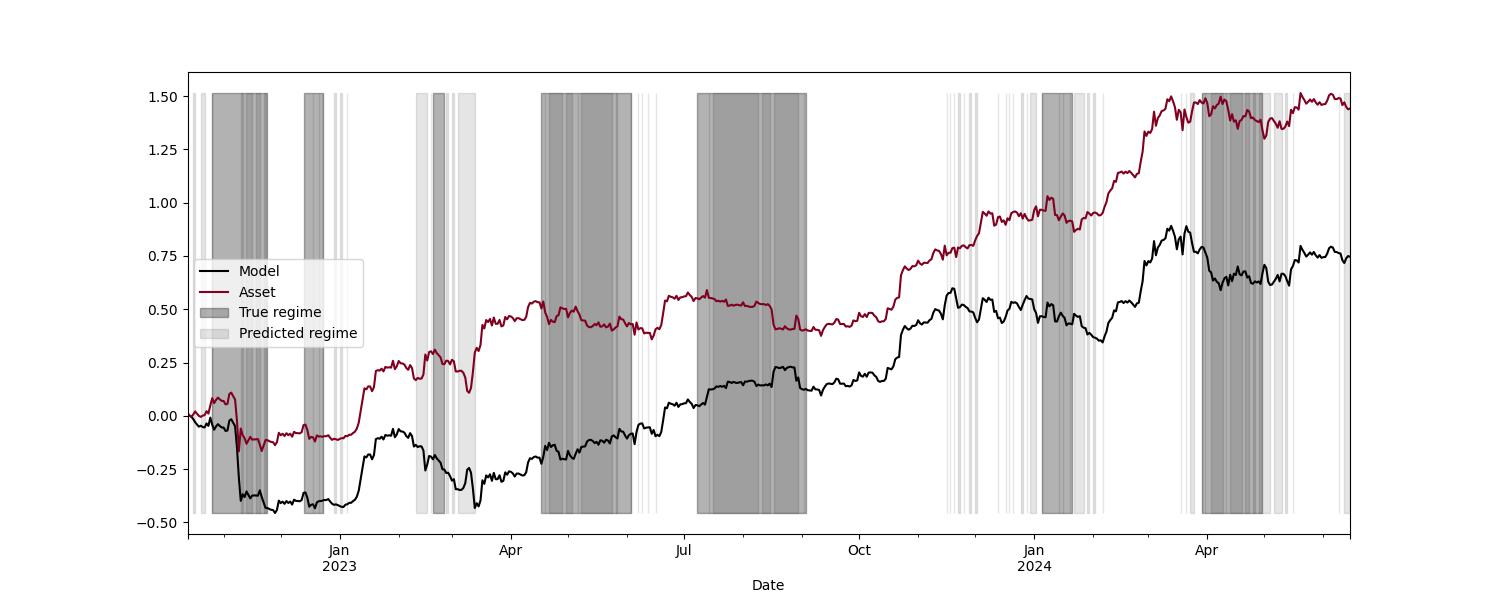}
\caption{m-TKAN (Out of sample)}
\label{fig:m-tkan-test}
\end{figure}

\end{document}